\documentclass[11pt]{article}
\usepackage{amsmath,amsfonts,amsmath,amssymb,amsthm}
\usepackage{comment}
\usepackage{bm}
\usepackage{graphicx,psfrag,epsf}
\usepackage{enumerate}
\usepackage{natbib}
\usepackage{algorithm}
\usepackage[noend]{algpseudocode}
\RequirePackage[colorlinks,citecolor=blue,linkcolor=blue,urlcolor=blue]{hyperref}

\usepackage{url} 
\newcommand{\blind}{0}

\usepackage{tikz}
\usepackage[capitalise]{cleveref}

\usepackage[shortlabels]{enumitem}

\newtheorem{theorem}{Theorem}

\newtheorem{assumption}{Assumption}

\newcommand{\mtry}{\texttt{mtry}}

\begin{document}

\def\spacingset#1{\renewcommand{\baselinestretch}%
{#1}\small\normalsize} \spacingset{1}

\if0\blind
{
  \title{\bf Getting Better from Worse:  \\  Augmented Bagging and a \\ Cautionary Tale of Variable Importance}
  \author{\Large{Lucas Mentch and Siyu Zhou} \\ ~ \\ Department of Statistics \\ University of Pittsburgh}

  \maketitle
} \fi

\if1\blind
{
  \bigskip
  \bigskip
  \bigskip
  \begin{center}
    {\LARGE\bf Getting Better from Worse:  Augmented Bagging and a Cautionary Tale of Variable Importance}
\end{center}
  \medskip
} \fi

\bigskip
\begin{abstract}
\noindent As the size, complexity, and availability of data continues to grow, scientists are increasingly relying upon black-box learning algorithms that can often provide accurate predictions with minimal \emph{a priori} model specifications.  Tools like random forests have an established track record of off-the-shelf success and even offer various strategies for analyzing the underlying relationships among variables. Here, motivated by recent insights into random forest behavior, we introduce the simple idea of augmented bagging (AugBagg), a procedure that operates in an identical fashion to classical bagging and random forests, but which operates on a larger, augmented space containing additional randomly generated noise features. Surprisingly, we demonstrate that this simple act of including extra noise variables in the model can lead to dramatic improvements in out-of-sample predictive accuracy, sometimes outperforming even an optimally tuned traditional random forest.  As a result, intuitive notions of variable importance based on improved model accuracy may be deeply flawed, as even purely random noise can routinely register as statistically significant.  Numerous demonstrations on both real and synthetic data are provided along with a proposed solution. \\
\end{abstract}


\spacingset{1.25}

\section{Introduction}
\label{sec:intro}

As data continues to become larger and more complex, scientists and analysts are increasingly relying upon adaptive learning methods in lieu of the more traditional parametric statistical models that require \emph{a priori} model specification.  Among these flexible alternatives, bagging \citep{Breiman1996} and random forests \citep{Breiman2001} have proven among the most popular and robust tools available with successful application in nearly every scientific field; for just a few select examples, see \cite{Diaz2006,Cutler2007,Bernard2007,Mehrmohamadi2016,Coleman2017}.  In a recent study, \cite{Fernandez2014} compared the performance of 179 classification methods across all datasets then available in the UCI Machine Learning Repository \citep{uci} and found random forests to be the top overall performer. In the previous two decades since their inception, numerous studies have sought to establish their important statistical properties including consistency \citep{Biau2010,Scornet2015,Klusowski2019sharp}, asymptotic normality \citep{Mentch2016,Wager2018}, and rates of convergence \citep{Peng2019} as well as means by which standard errors \citep{Sexton2009}, confidence intervals \citep{Wager2014,Mentch2016}, and hypothesis testing procedures \citep{Mentch2016,Mentch2017,Coleman2019} can be obtained.

Bagging, first introduced in a tree-based setting by \cite{Breiman1996}, involves drawing $B$ bootstrap samples from the original training data, refitting the base model (tree) on each, and averaging the individual outputs to obtain the final predictions.  When base models are traditional classification or regression trees \citep{CART}, at each internal node, the optimal empirical split point is chosen by searching over all features and potential splits. Random forests can thus be seen as a less-greedy alternative, whereby eligible features for splitting are randomly selected at each internal node.  

Despite the abundance of forest-related work in recent years, substantially less effort has been devoted to principled studies of the inner workings of random forests that might more fully explain their robust record of success.  Recently however, \cite{Mentch2019} suggested that the additional randomness utilized in random forests was simply an implicit form of regularization.  The $\mtry$ parameter in random forests that dictates the number of available features at each split could therefore been seen as akin to the $\lambda$ shrinkage penalty in explicit regularization methods like ridge regression \citep{Hoerl1970} and lasso \citep{Tibshirani1996}.  \cite{Mentch2019} suggested that the random subsampling of features helped the trees to avoid overfitting and that this was particularly beneficial in low signal-to-noise ratio settings.  \cite{Lejeune2019} demonstrated a similar effect for ensembles consisting of linear model base learners fit via ordinary least squares (OLS).  

The idea that the randomness in random forests serves as a means of regularization not only eliminates some of the mystery of their sustained success but also suggests that alternative modifications to the standard bagging procedure that also induce some means of regularization may produce similar gains in accuracy.  In this work, we propose one such alternative we refer to as \emph{augmented bagging} (AugBagg) wherein the original feature space is augmented with additional noise features generated conditionally independent of the response, after which the standard bagging procedure is carried out.  Alarmingly, in many instances, we show that this simple act of adding extra random noise features to the model can greatly improve its out-of-sample predictive accuracy.  In fact, when applied to the worst-tuned model, this counterintuitive action can transform it into one that is more accurate than even the best-tuned model on the original data.  Rather than making a bad model worse as most would presume, the addition of otherwise useless random noise features can have precisely the opposite effect.  Importantly, this finding is not merely the byproduct of the mysterious inner workings of random forests.  Very recent work by \cite{kobak2019optimal} has shown that including additional random noise features in the regression can also improve the performance of linear models, albeit under a setup that is a bit more strict.

This finding has crucial implications for the ways in which we measure and test feature importance.  In black-box contexts where traditional measures like p-values may be unavailable or difficult to obtain, numerous recent studies have formally proposed methods to evaluate feature importance by measuring the change in accuracy when the features of interest are dropped from the model \citep{Mentch2016,Mentch2017,Lei2018,Coleman2019,Williamson2020}.  The implicit logic in such procedures feels intuitive and obvious: if the response can be more accurately predicted when a supplemental collection of features are included in the model, then those additional features must hold some information about the response beyond whatever is contained in the original features.  This work, however, demonstrates that this widely-held belief is simply false.  In many instances, independent random noise can itself improve predictions, thereby leading to seemingly paradoxical situations in which features that are independent of the response routinely register as statistically significant.  Much further discussion on these implications is included in the latter sections of this work along with a proposed solution.

The remainder of this paper is laid out as follows.  In Section \ref{sec:augbagg} we formally introduce the AugBagg procedure and in Section \ref{sec:sims} we provide numerous simulations and real-data experiments to demonstrate its surprisingly competitive predictive performance.  In Section \ref{sec:theory} we provide theoretical motivation for the AugBagg procedure, building upon very recent results established for other learning procedures.  Implications for measuring and testing variable importance are discussed in Section \ref{sec:varimp}, where we also suggest a more robust alternative testing framework in which tests for feature importance maintain the nominal level for noise features, even when such features are capable of producing non-trivial gains in accuracy.

\section{Augmented Bagging}
\label{sec:augbagg}
Throughout the remainder of this paper, we assume data of the form $\mathcal{D}_n = \{\bm{Z}_1, ..., \bm{Z}_n\}$ where each ordered pair $\bm{Z}_i = (\bm{X}_i,Y_i)$ consists of a feature vector $\bm{X}_i = (X_{1,i}, ..., X_{p,i})$ and response $Y_i \in \mathbb{R}$.  Given $B$ bootstrap samples of the data, the original bagging procedure \citep{Breiman1996} generates a prediction at $\bm{x}$ of the form
\begin{equation}
\hat{y}_{\text{Bagg}} = \frac{1}{B} \sum_{b=1}^{B} T(\bm{x}; \, \omega_b, \mathcal{D}_n)
\label{eqn:bagg}
\end{equation}
where the randomness $\omega_b$ serves only to select the bootstrap sample on which the $b^{th}$ tree $T$ is trained.  Whenever the randomness is assumed to select both the bootstrap sample as well as the $\mtry < p$ eligible features at each internal node, the random forest prediction $\hat{y}_{\text{RF}}$ can be written in the same general form as (\ref{eqn:bagg}).

The augmented bagging (AugBagg) procedure we introduce here represents a straightforward extension of classical bagging.  Beginning with the original dataset $\mathcal{D}_n$, we create an augmented dataset $\mathcal{D}_{n}^{*}$ consisting of additional noise features generated conditionally independent of $Y$.  This augmented dataset thus takes the form $\mathcal{D}_{n}^{*} = \{\bm{Z}_{1}^{*}, ..., \bm{Z}_{n}^{*}\}$ where each $\bm{Z}_{i}^{*}$ now denotes an ordered triplet $(\bm{X}_i,\bm{N}_i,Y_i)$ consisting of the original features $\bm{X}_i$ and response $Y_i$, but also an additional set of noise features $\bm{N}_i = (N_{1,i}, ..., N_{q,i})$. The original bagging procedure is then performed on this augmented feature space so that the AugBagg output produces predictions of the form
\begin{equation}
\hat{y}_{\text{AugBagg}} = \frac{1}{B} \sum_{b=1}^{B} T((\bm{x,n}); \, \omega_b, \mathcal{D}_{n}^{*})
\label{eqn:AugBagg}
\end{equation}
\noindent where $\bm{n}$ can be filled in with random draws from the additional noise features.

Importantly, we insist only that $\bm{N}$ be generated conditionally independent of $Y$ given $\bm{X}$.  This thus allows for additional noise features to be correlated with the original features.  As demonstrated in the following sections, the manner in which noise features are generated can greatly impact performance.

\section{Simulations and Real Data Examples}
\label{sec:sims}

We now present a number of simulation studies to demonstrate the effectiveness of the AugBagg procedure in practice.  To begin, we consider a standard linear model of the form $Y = \bm{X}\beta + \epsilon$ with $n \times p$ design matrix, the rows of which are i.i.d.\ multivariate normal $\mathcal{N}_p(\bm{0},\Sigma)$ where $\Sigma \in \mathbb{R}^{n \times p}$ has entry $(i, j) = \rho^{|i-j|}$ with $\rho = 0.35$.  The form of this covariance corresponds to that utilized frequently in the recent work by \cite{Mentch2019} and to the `beta-type 2' setup utilized in \cite{Hastie2017}.  The original data includes $n=100$ observations, $p=5$ original signal features with $\beta_1 = \cdots = \beta_5 = 1$, and $q$ additional i.i.d.\ noise features sampled from $\mathcal{N}(0,1)$ independent of $\bm{X}$ are then added with $q$ ranging from 1 to 250.  As in \cite{Mentch2019} and \cite{Hastie2017}, the noise term $\epsilon$ is sampled from $\mathcal{N}(0,\sigma^{2}_{\epsilon})$ with $\sigma_{\epsilon}^{2}$ chosen to satisfy a particular signal-to-noise ratio (SNR), given in this context by $\beta^{T}\Sigma\beta / \sigma_{\epsilon}^{2}$.

\begin{figure}[t!]
	\centering
	\includegraphics[width =0.8\linewidth]{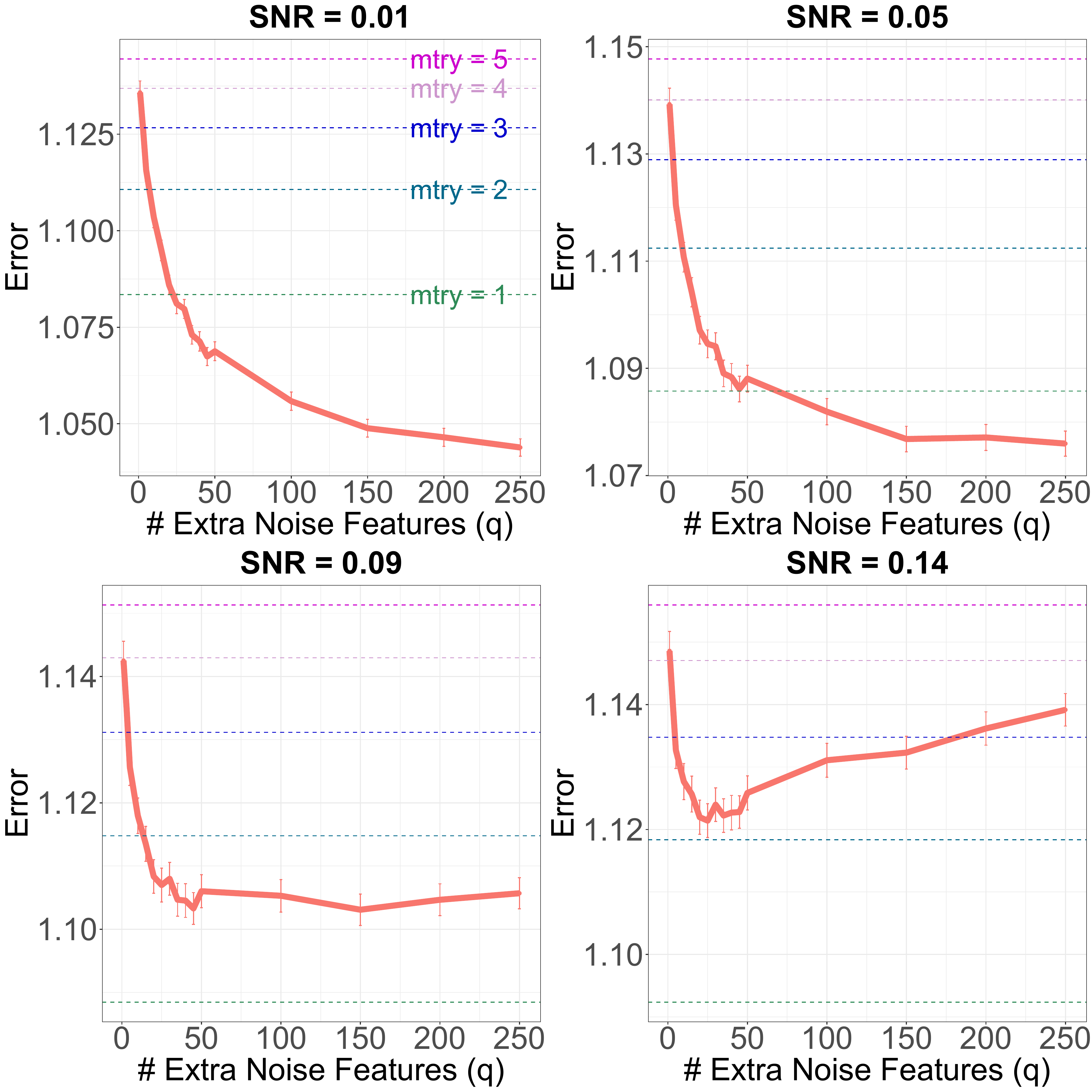}
	\caption{Performance of Augmented Bagging as $q$ additional independent noise features are added to the model as compared with random forests and traditional bagging ($\mtry=5$) built on the original data.  Each point in each plot corresponds to the average error after repeating the experiment 500 times with error bars showing $\pm1$ standard deviation.}
	\label{fig:AugBagg1}
\end{figure}

Figure \ref{fig:AugBagg1} shows the performance of the AugBagg procedure against that of traditional bagging and random forests that are built on the original data consisting of only the first $p=5$ features.  Each plot corresponds to a different SNR (0.01, 0.05, 0.09, or 0.14) and shows the relative test error, defined as the test MSE calculated on an independent, randomly generated test set of 1000 observations, scaled by $\sigma^{2}_{\epsilon}$.  Each point in each plot corresponds to the error averaged over 500 iterations with error bars showing $\pm1$ standard deviation.  Horizontal lines in each plot correspond to the accuracy of a random forest constructed with a particular value of $\mtry$ and we see that in each case the random forest error grows as $\mtry$ increases.  At the lowest SNR of 0.01, AugBagg appears to continually improve with $q$, easily surpassing the best random forest at around $q=25$.  The results are similar, though less dramatic, when the SNR is increased to 0.05.  When the SNR is increased to 0.09, the performance of AugBagg appears to level-off around $q=50$, never achieving that of the optimal random forest with $\mtry=1$.  Finally, when the SNR is 0.14, the additional noise features appear to help until approximately $q=50$, after which point the performance begins to deteriorate.

\begin{figure}
	\centering
	\includegraphics[width =0.32\linewidth]{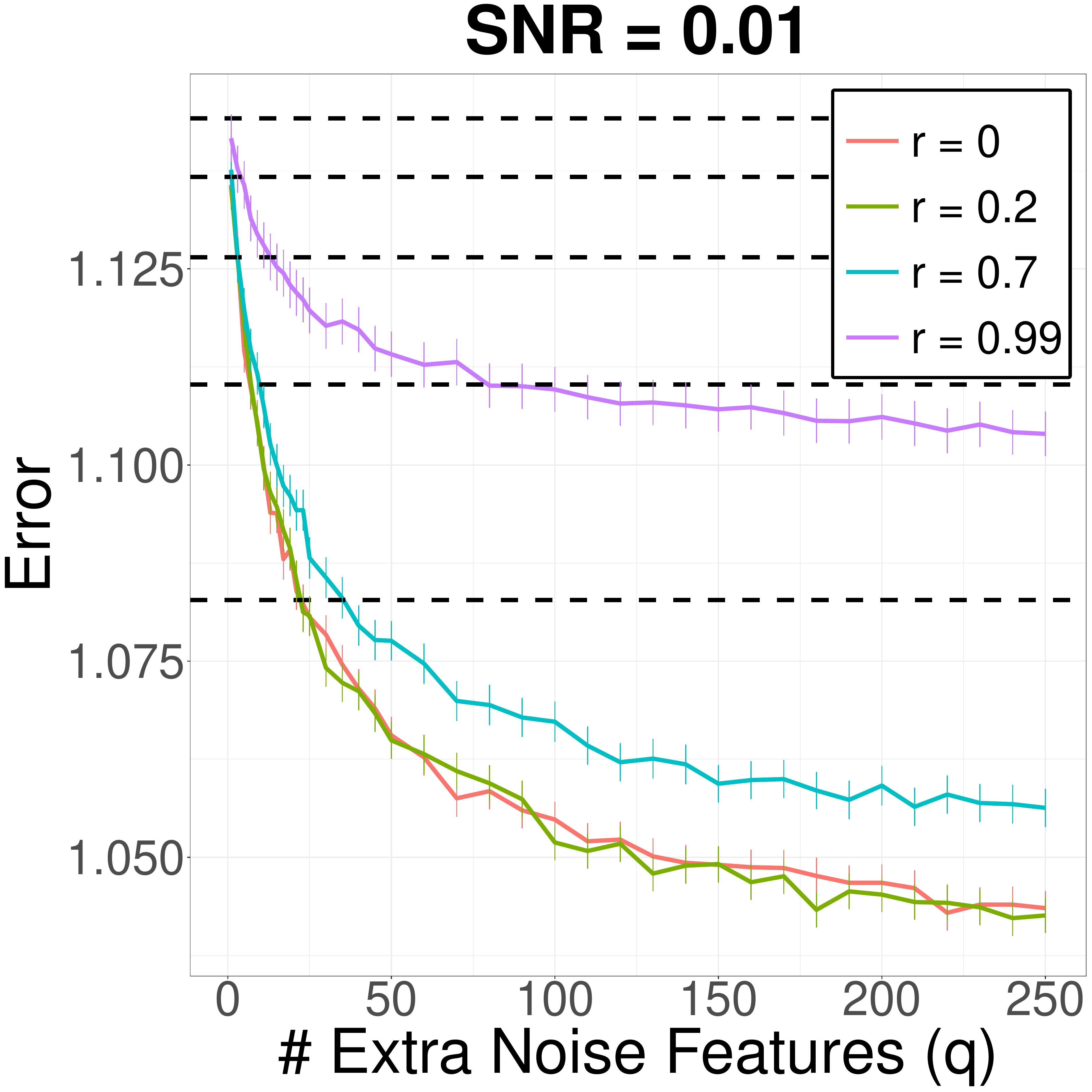}
	\includegraphics[width =0.32\linewidth]{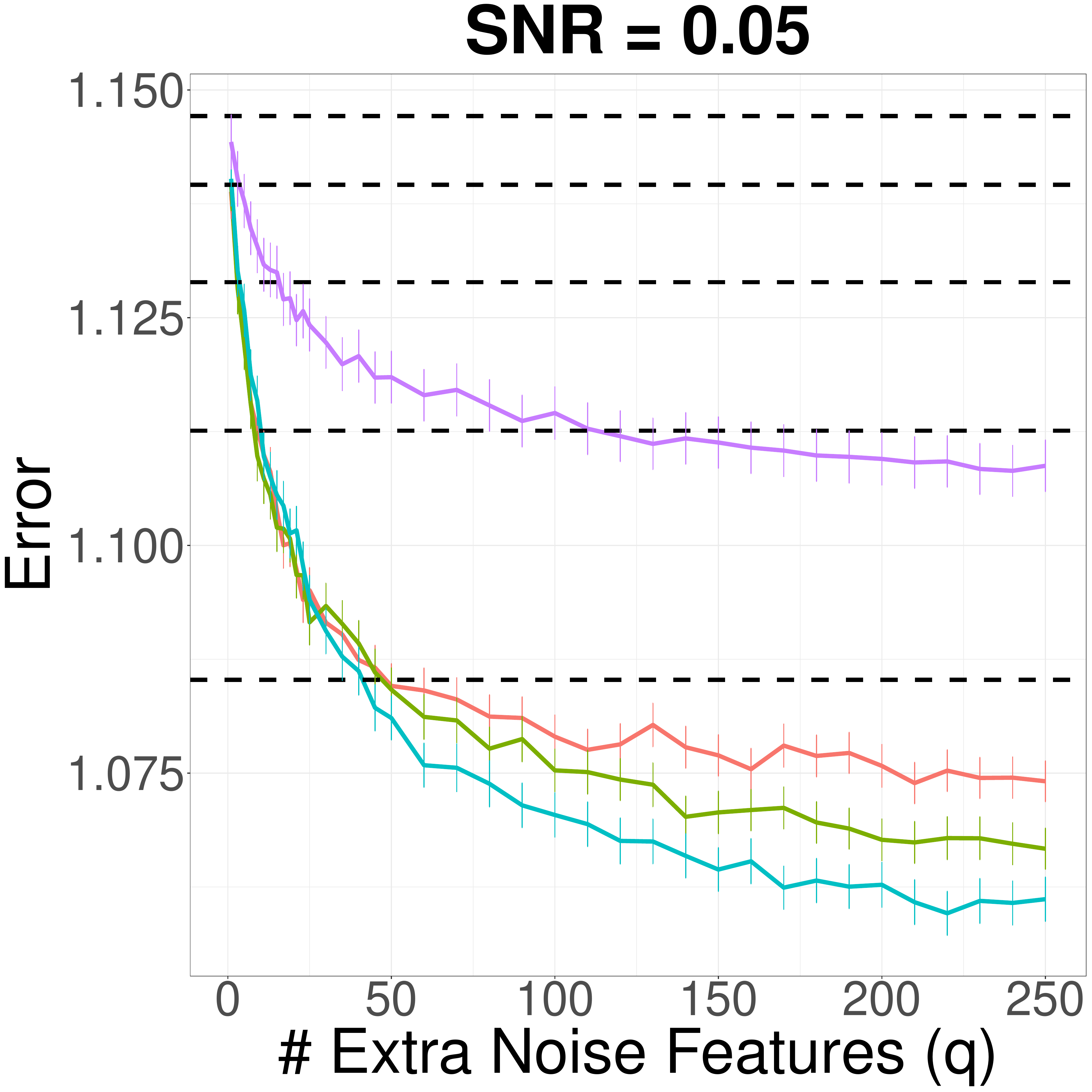}
	\includegraphics[width =0.32\linewidth]{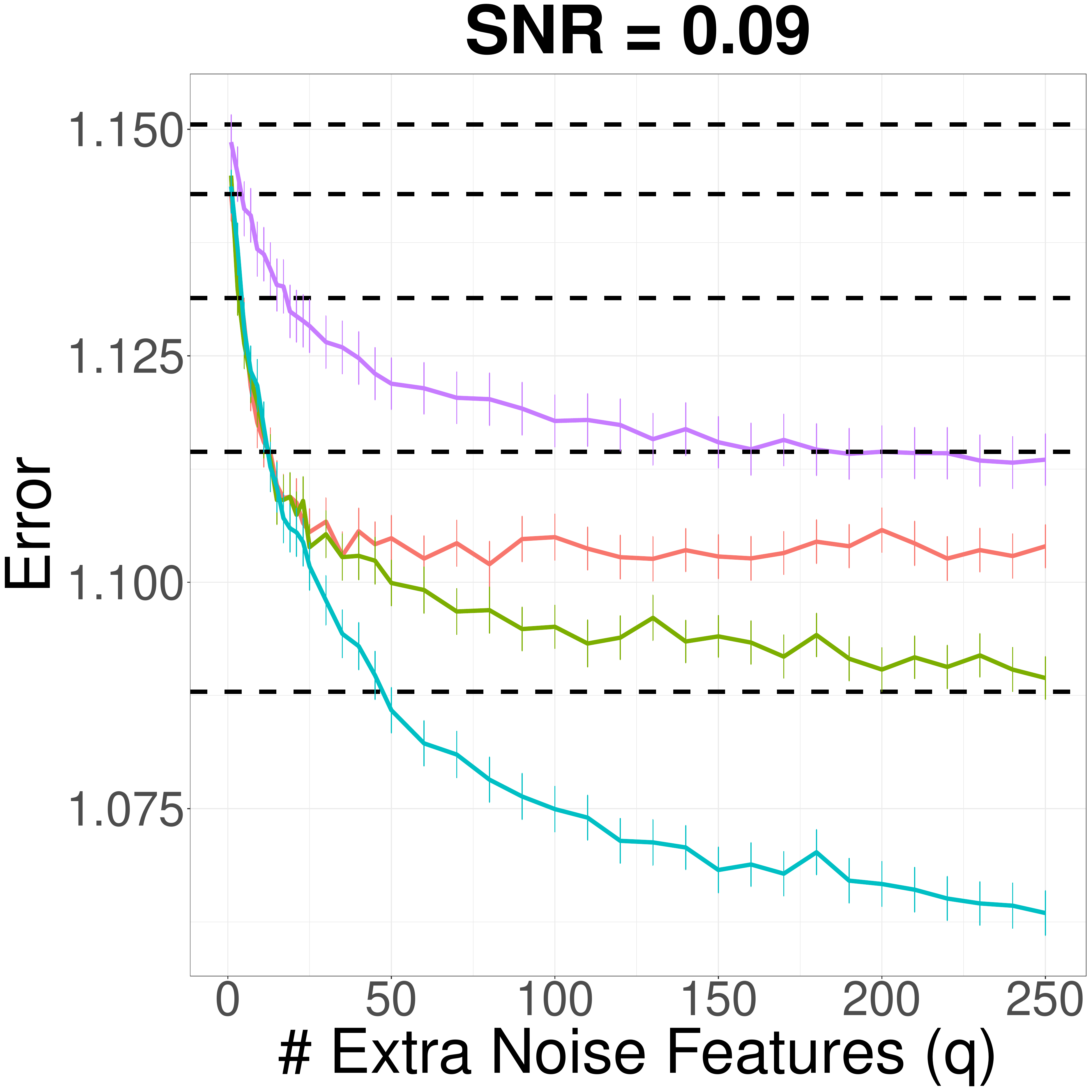} \\
	\vspace{2mm}
	\includegraphics[width =0.32\linewidth]{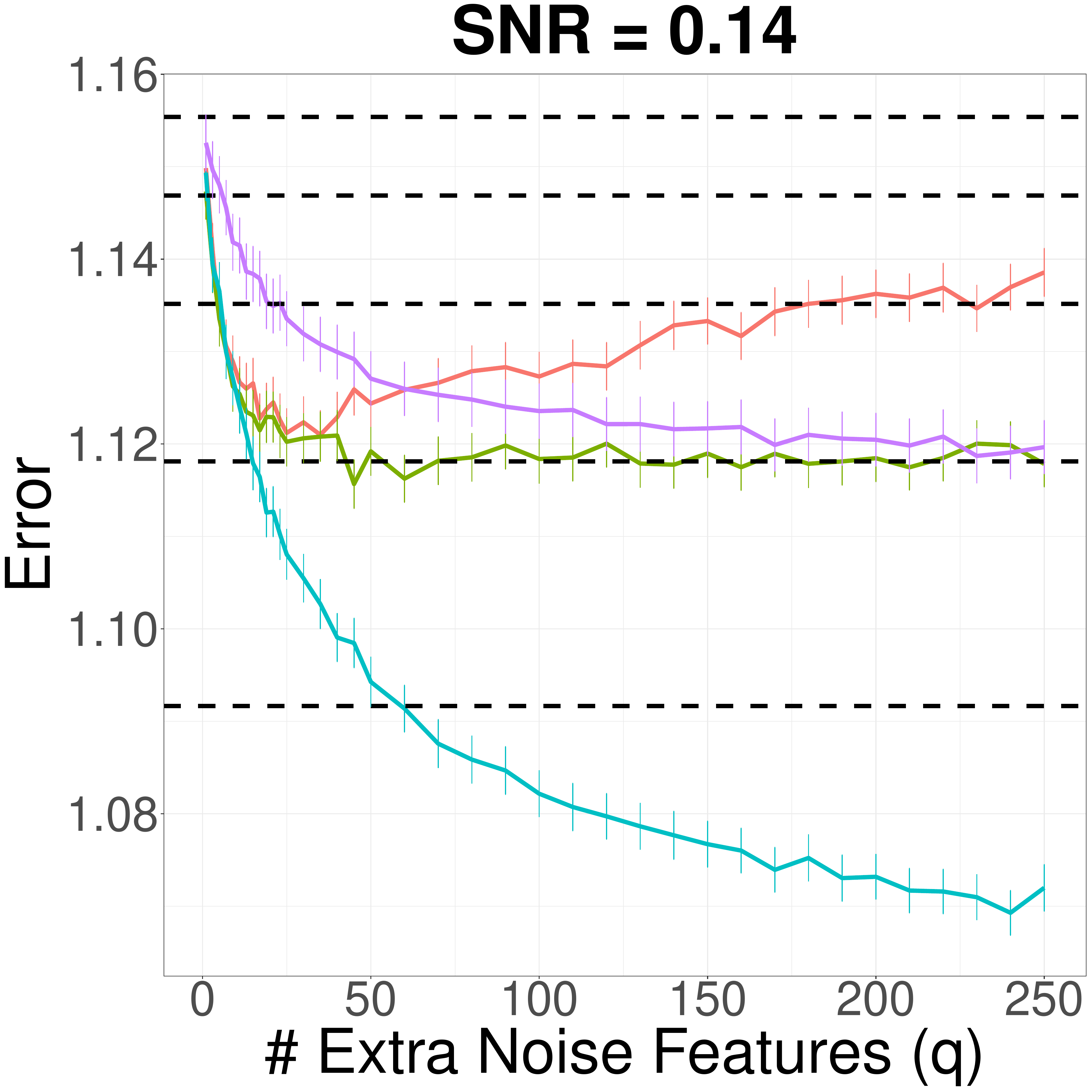}
	\includegraphics[width =0.32\linewidth]{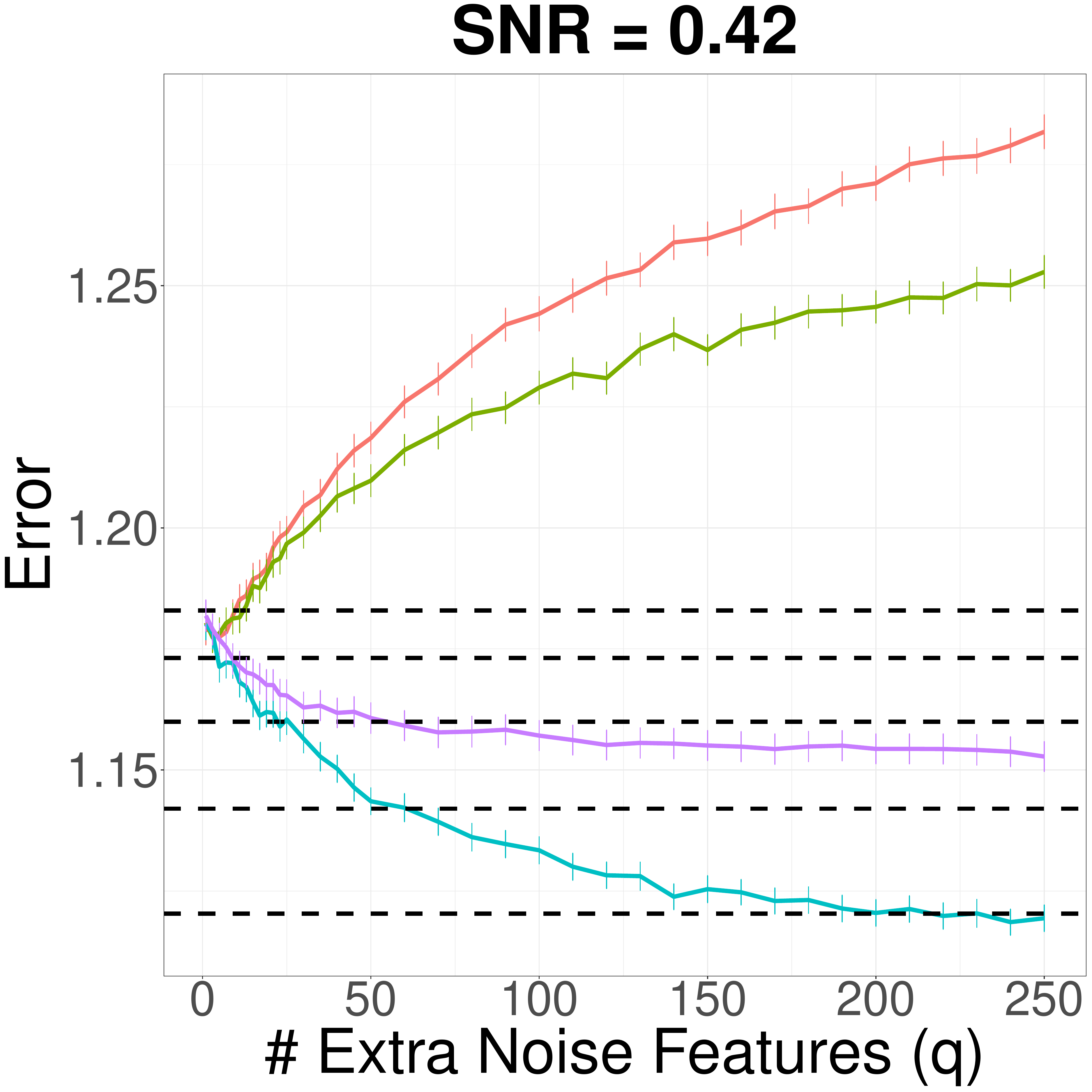}
	\includegraphics[width =0.32\linewidth]{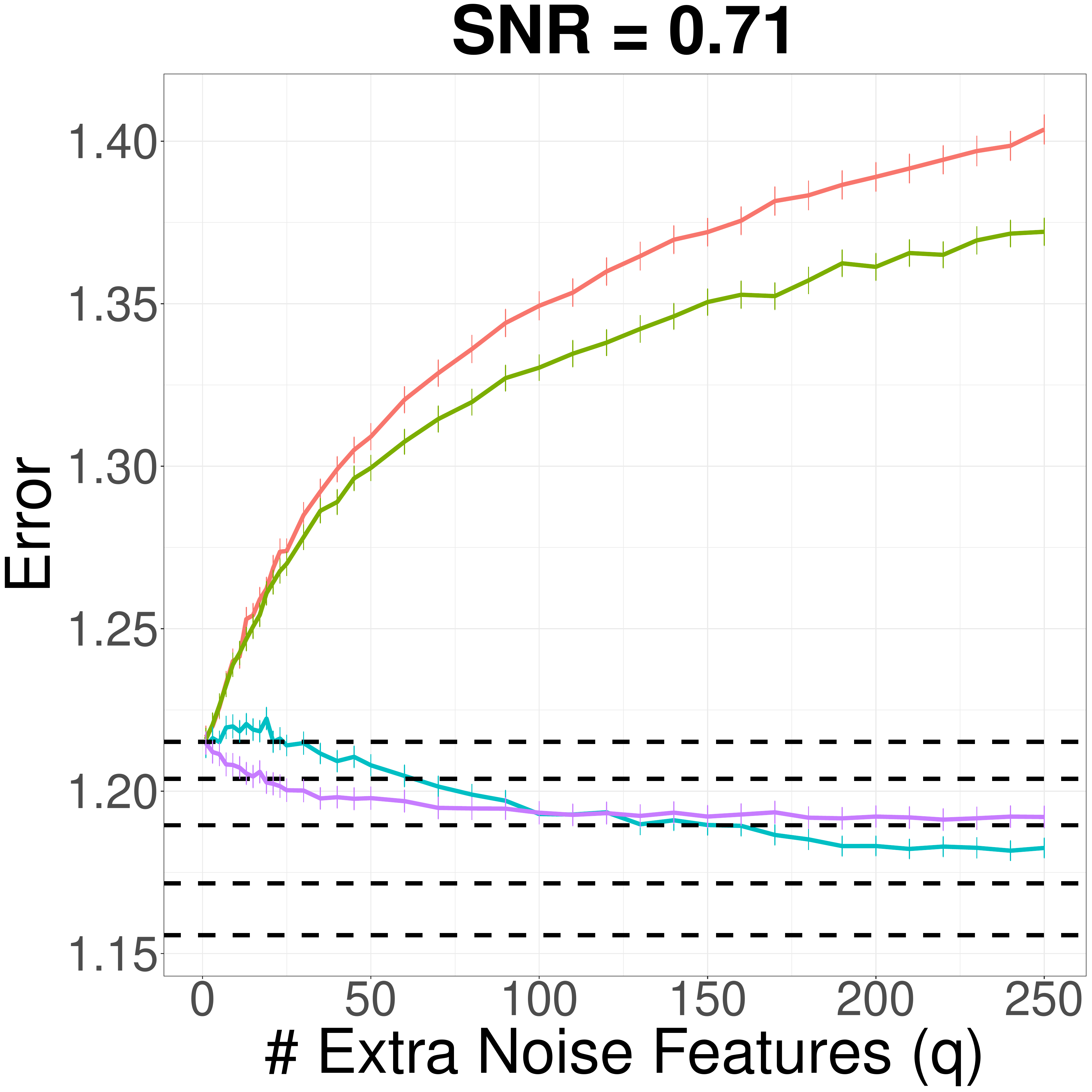} \\
	\vspace{2mm}
	\includegraphics[width =0.32\linewidth]{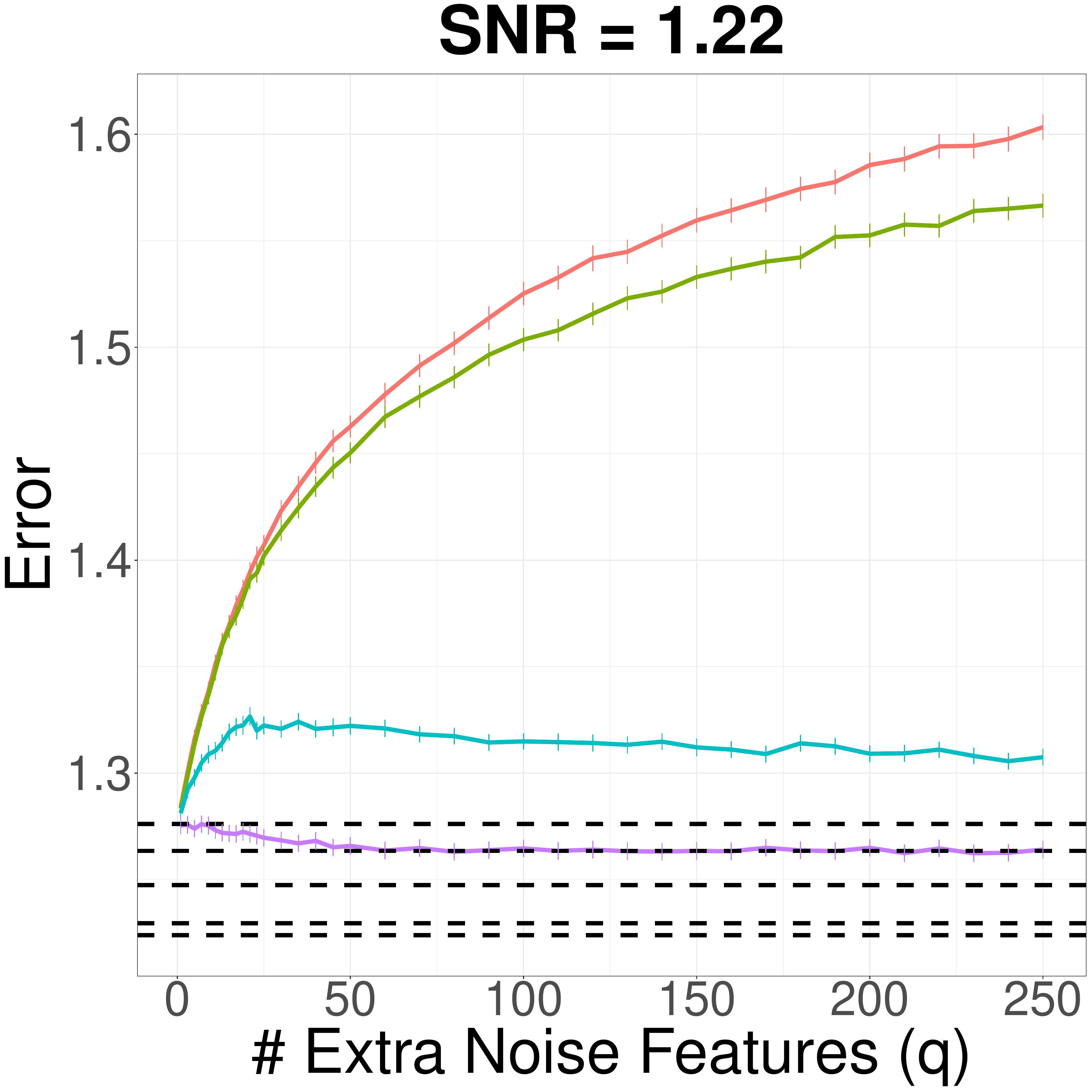}
	\hspace{7mm}
	\includegraphics[width =0.32\linewidth]{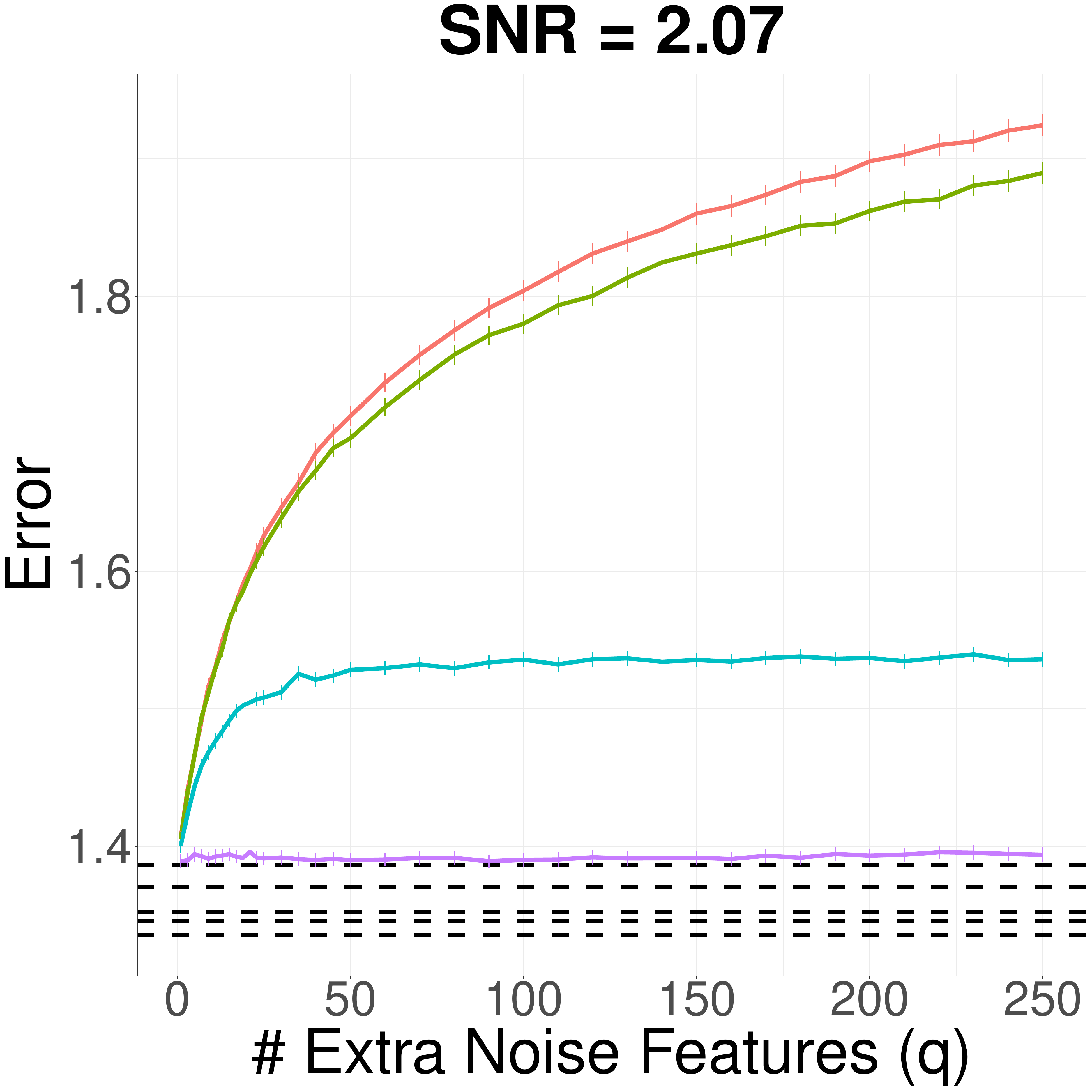}
	\caption{Performance of AugBagg compared against random forests as additional noise variables are added to the model.  Different colored lines in each plot correspond to different correlation strengths between the original and noisy additional features.}
	\label{fig:AugBagg2}
\end{figure}

The results in Figure \ref{fig:AugBagg2} expand these simulations.  The data and model setup remain the same but the results are explored over a wider range of 8 SNRs ranging from 0.01 to 2.07.  Furthermore, in addition to the additional noise features sampled independently of $\bm{X}$, here we consider the addition of noisy features that are correlated with one of the first 5 signal features.  In a similar fashion to knockoffs \citep{Barber2015,Candes2018}, such noise features are thus independent of the response $Y$ given $\bm{X}$.  To generate such features, we first select an original feature $X$ at random and generate a standard normal $Z \sim \mathcal{N}(0,1)$.  For a given level of correlation $r$, each of the additional $q$ features then take the form
\begin{equation}
N = rX + \sqrt{(1-r^2)} Z.
\end{equation}
\noindent In each of the plots in Figure \ref{fig:AugBagg2} we consider correlations of $r = 0, 0.2, 0.7$, and 0.99 for batches of additional features ranging in size from 1 to 250. Performance is measured in the same fashion and estimates are averaged over 500 replications for each point in each plot.  In the following discussion, we will use the shorthand $AB(q,r)$ to denote an AugBagg model with $q$ additional noise features, each of which has correlation $r$ with one of the features in the original dataset.

Figure \ref{fig:AugBagg2} presents a very interesting and telling story in terms of how the additional noise features are influencing performance and how that influence changes across different SNR levels.  Looking only at Figure \ref{fig:AugBagg1} where the noise features are independent of both the response and the original features, one might suspect that this phenomenon occurs only at very low SNRs.  Looking at Figure \ref{fig:AugBagg2} however, we see that when the noise features are correlated with the original features, improvements in model accuracy are seen even at relatively high SNRs.

At the lowest SNRs of 0.01 and 0.05, we see that in every case, the AugBagg models are improving with the number of extra noise features $q$.  Once $q>50$, all AugBagg models begin to outperform even the best random forest, with the exception of $AB(q,0.99)$ where very highly correlated noise features are added.  At SNR = 0.09, much the same story is present but now only $AB(q,0.7)$ outperforms the optimal random forest and again this transition happens around $q=50$.  At SNR = 0.14 we begin to see an interesting shift where the performance of the independent noise model $AB(q,0)$ begins to deteriorate with $q$.  When the SNR grows to 0.42 and 0.71, this effect is much more pronounced with $AB(q,0)$ and $AB(q,0.2)$ both deteriorating with $q$.  At the largest SNRs of 1.22 and 2.07, $AB(q,0.99)$ is now the only model not deteriorating substantially with $q$.

\subsection{Experiments on Real World Data}
\label{subsec:realdata}
The previous simulations demonstrate that the AugBagg procedure can lead to substantial gains in accuracy over the baseline bagging procedure on synthetic datasets.  Following a very similar setup to \cite{Mentch2019}, we now investigate its performance on a variety of real-world datasets.  Data summaries are provided in Table \ref{tab:uci}; a total of nine low-dimensional ($p<n$) and five high-dimensional ($p>n$) datasets are included.

In implementing the AugBagg procedure, we consider tuning both the number of additional noise features $q$ as well as the level of correlation $r$.  Since different datasets have different numbers of original features, $q$ is tuned over $p/2$, $p$, $3p/2$ and $2p$. The correlation strength $r$ is tuned over 0, 0.1, 0.4, 0.7 and 0.9.  In datasets with mixed feature types, each additional noise feature is chosen to be correlated with one randomly selected continuous feature from the original data. As in \cite{Mentch2019}, we inject further noise of the form $\epsilon \sim N(0,\sigma_{\epsilon}^{2})$ into the response where the variance of the noise $\sigma^{2}_{\epsilon}$ is chosen as some proportion of $\hat{\sigma}^{2}_{y}$, the estimated variance of the original response $Y$. Performance is measured in terms of relative test error (RTE), defined as
\begin{align}
	\text{RTE} = \frac{\widehat{Err}(bagging) - \widehat{Err}(AugBag)}{\hat{\sigma}^2_y} \times 100\%
\end{align}
\noindent with positive values indicating superior performance by AugBagg. 


\begin{figure}[t!]
	\centering
	\includegraphics[width=0.47\textwidth]{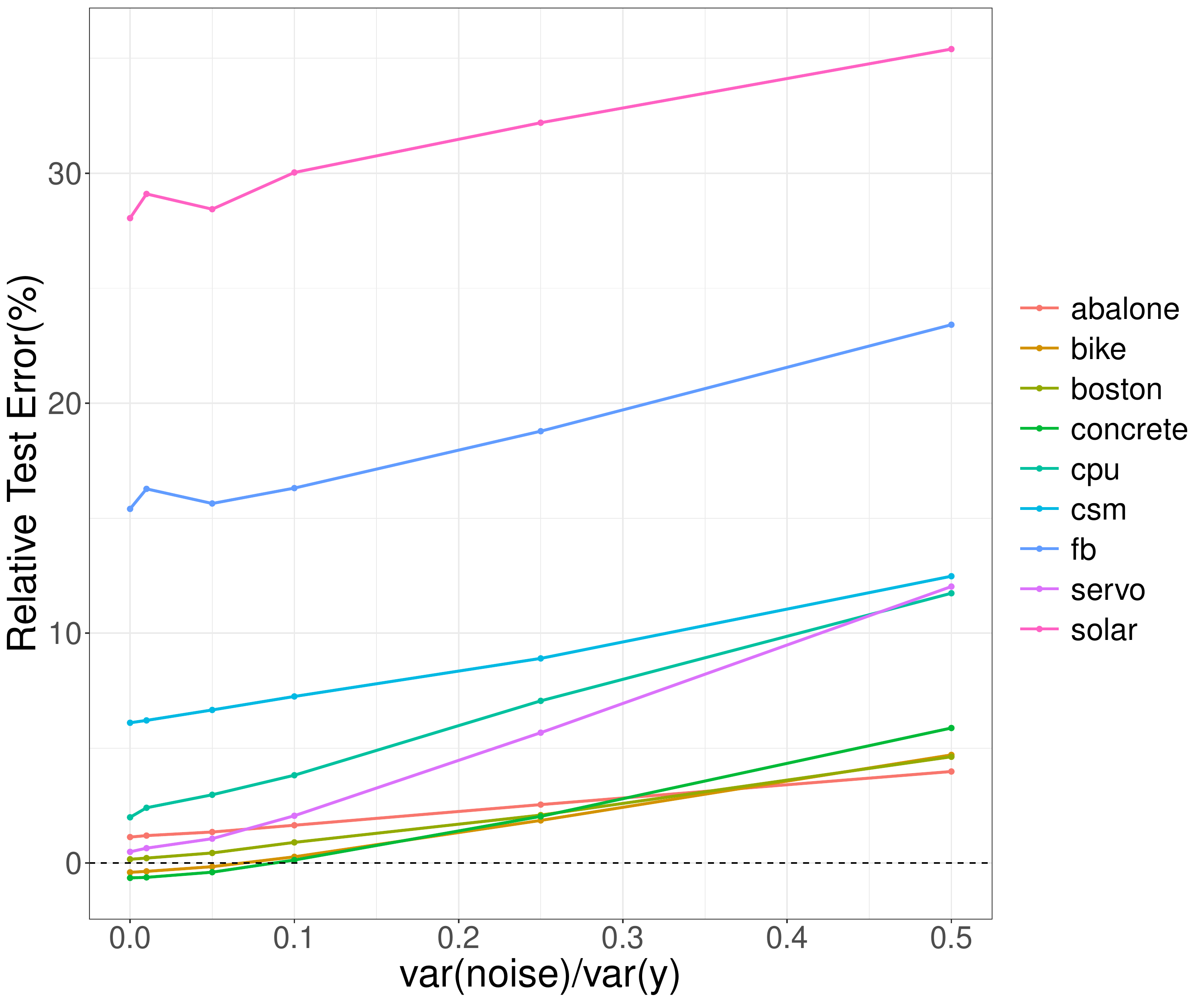}
	\includegraphics[width=0.47\textwidth]{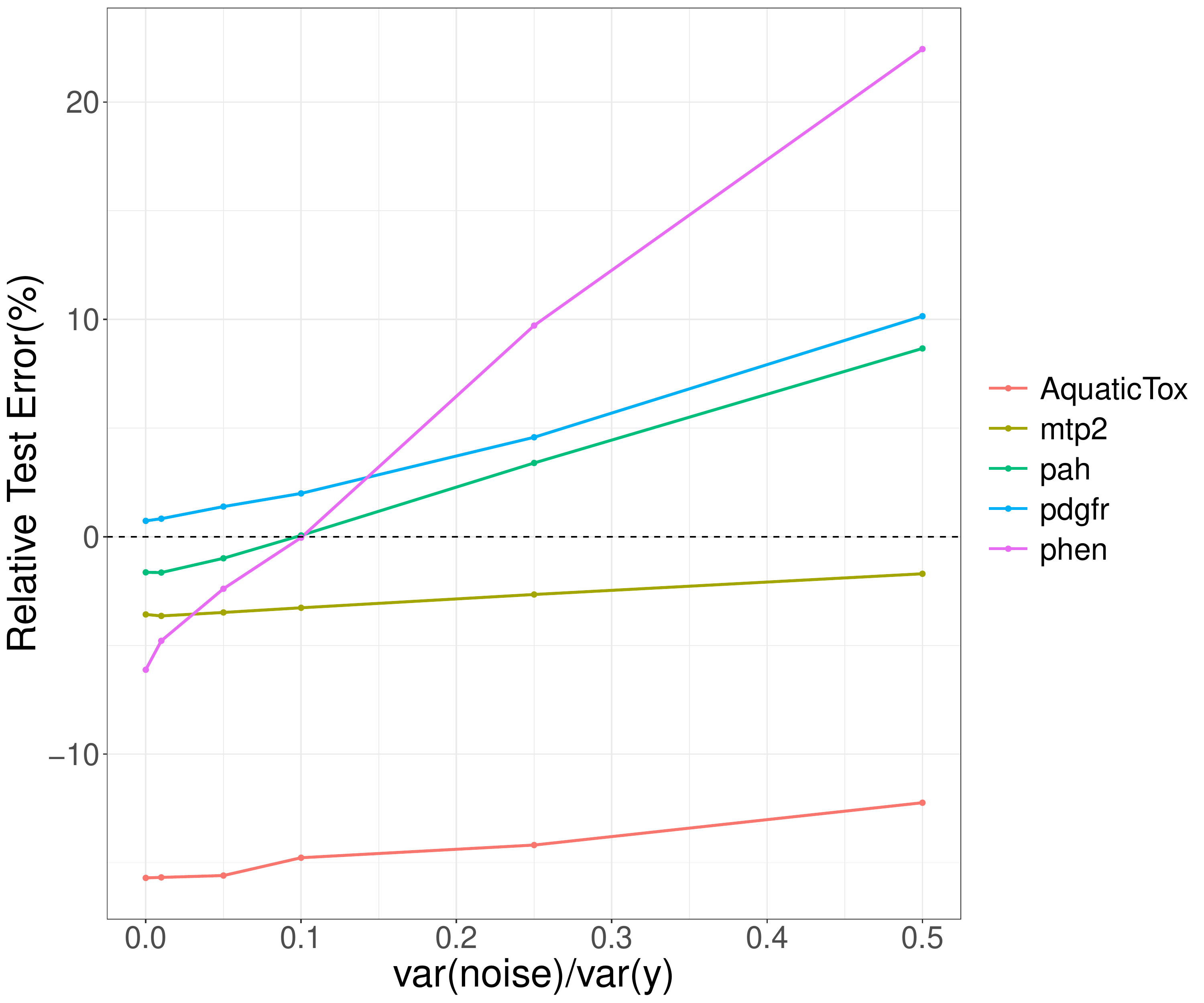}
	\caption{Relative test error (RTE) on real datasets with additional noise added onto the response. Left: low-dimensional datasets. Right: high-dimensional datasets.}
	\label{fig:AugBag_RealDt_RTE}
\end{figure}

\begin{table}[t]
	\centering
	\begin{tabular}{lcc}
	\hline
	Dataset	&	$p$	&	 $n$		\\
	\hline
	Abalone Age [\texttt{abalone}] \cite{UCIabalone}	&	8			&	4177		\\
	Bike Sharing [\texttt{bike}] \cite{UCIbike}	&	11			&	731		\\
	Bioston Housing [\texttt{boston}]	\cite{UCIboston} &	13			&	506		\\
	Concrete Compressive Strength [\texttt{concrete}] \cite{UCIconcrete}	&	8			&	1030		\\
	CPU Performance [\texttt{cpu}] \cite{UCIcpu}	&	7			&	209		\\
	Conventional and Social Movie [\texttt{csm}] \cite{UCIcsm}	&	10			&	187		\\
	Facebook Metrics [\texttt{fb}] \cite{UCIfb}	&	7			&	499		\\
	Servo System [\texttt{servo}] \cite{UCIservo}	&	4			&	167		\\
	Solar Flare [\texttt{solar}] \cite{UCIflare} &	10			&	1066		\\
	Aquatic Toxicity [\texttt{AquaticTox}] \cite{he2005assessing}	&	468	&	322	\\
	Molecular Descriptor Influencing Melting Point [\texttt{mtp2}] \cite{bergstrom2003molecular}	&	1142			&	274		\\
	Weighted Holistic Invariant Molecular Descriptor [\texttt{pah}] \cite{todeschini1995weighted}	&	112			&	80 		\\
	Adrenergic Blocking Potencies [\texttt{phen}] \cite{cammarata1972interrelation}	&	110			&	22			\\
	PDGFR Inhibitor [\texttt{pdgfr}] \cite{guha2004development}	&	320			&	79			\\
	\hline
	\end{tabular}
	\caption{Summary of datasets utilized. }
	\label{tab:uci}
\end{table}

Results are shown in Figure \ref{fig:AugBag_RealDt_RTE}. In every case, the performance of the tuned AugBagg procedure increases as more noise is added to the response, as demonstrated by the positive slope displayed for each dataset.  In 12 of the 14 datasets, AugBagg quickly begins to outperform bagging on the original data with substantial improvements occurring as more noise is injected.  Furthermore, it is interesting to note that in the two cases where traditional bagging remains superior, both datasets (\texttt{AquaticTox} and \texttt{mtp2}) are high-dimensional and, in fact, contain the largest number of original features out of all datasets considered ($p=468$ and 1142, respectively).  In these cases, it is quite possible that many of the original features are themselves noisy and thus the additions we make are of no further benefit.  Indeed, an optimally tuned lasso model built on the \texttt{AquaticTox} and \texttt{mtp2} datasets selects only (approximately) 17\% and 4\% of the features, respectively.

\section{Theoretical Motivation and Analogous Results}
\label{sec:theory}

In the following three subsections, we draw upon recent results on interpolation and implicit regularization in order to provide some theoretical motivation for the practical success of the AugBagg procedure.

\subsection{Randomization as Regularization}
In very recent work, \cite{Mentch2019} argue that the success of random forests is due in large part to a kind of implicit regularization offered by the \texttt{mtry} parameter governing the number of features available for splitting at each node.  The authors demonstrate that the lower the signal-to-noise ratio (SNR) of the data, the smaller the optimal value of \texttt{mtry}.  Moreover, the authors demonstrate that this behavior is not tree-specific, but holds for any ensemble consisting of forward-selection-style base learners in which the available features are randomly restricted at each step.  Specifically, the authors consider a type of randomized forward selection (RandFS) that proceeds in the same fashion as a standard linear model forward selection process, but where only a randomly selected subset of the remaining features are eligible to be added to the model at each step. 

Given data of the form described above, consider a generic regression relationship of the form $Y = f(\bm{X}) + \epsilon$ and consider an estimate $\hat{f}_{RFS}$ formed by averaging over $B$ individual RandFS models $\hat{f}_{RFS,1}, ..., \hat{f}_{RFS,B}$.  Each of the individual RandFS models can be written as 
\[
\hat{\beta}^{(b)}_{\text{RFS}} = \hat{\beta}_{0}^{(b)} + X_{(1)}^{(b)} \hat{\beta}_{(1)}^{(b)} + \cdots + X_{(d)}^{(b)} \hat{\beta}_{(d)}^{(b)}
\]
\noindent where $X_{(j)}^{(b)}$ is the feature selected at the $j^{th}$ step in the $b^{th}$ model and $\hat{\beta}_{(j)}^{(b)}$ is the corresponding coefficient estimate. Given an orthogonal design matrix, the authors note that for any given feature $X_j$, the corresponding coefficient estimate in each model is either 0 if $X_j$ is not included in the model or it equals the ordinary least squares estimate $\hat{\beta}_{j,OLS}$ if $X_j$ is selected.  Averaging across $B$ models of this form thus yields a term in the final model of the form $\gamma_j \cdot \hat{\beta}_{j,OLS}$ where $\gamma_j$ corresponds to the proportion of individual RandFS models in which $X_j$ was included.  

In this sense, the ensemblized RandFS procedure can be seen as producing shrinkage and the amount of shrinkage $\gamma_j$ on each feature depends on both the probability that the feature is made eligible and the probability that the feature is actually selected if made available.  While the later probability depends on the particular modeling technique and loss function employed, the probability of being made eligible is a direct function of only \texttt{mtry}.  

But the previous statement is only valid under the typical ``fixed $p$" setup where the dimensionality of the feature space is assumed fixed.  Suppose instead that \texttt{mtry} is held fixed and that the procedure is repeated on an augmented feature space where more noise variables are added.  Then under the same setup as above, it's clear that $\gamma_j$ decreases as a function of the number of extra noise features $q$ since each original feature will thus have a lower probability of being made eligible.  However, even for large values of $\mtry$, we argue further that the probability of being selected once eligible also decreases as $q$ increases and that such a decrease can be particularly dramatic for features only weakly related to the response.  Put simply, for a given original feature $X_j$, as more noise features are added to the model, the probability that some of those new features will appear at least as important as $X_j$ grows with $q$.  Thus, even for large values of $\mtry$ where the procedure begins to resemble that of bagging, the augmented version of the procedure may produce a similar kind of regularization and shrinkage to that offered by traditional random forests.

\subsection{AugBagg and OLS Ensembles}
While the recent work of \cite{Mentch2019} utilized linear model forward selection settings in order to better illustrate the regularization effect of random forests, in work appearing around the same time, \cite{Lejeune2019} provided an in-depth analysis focused on ensembles where each base learner is simply a linear model constructed on a subsample of features and observations with coefficients estimated via ordinary least squares.  As in \cite{Mentch2019}, the authors observe that feature subsampling at the base-learner stage produces a regularization effect, concluding that for optimally-tuned subsampling rates, the asymptotic risk of the OLS ensemble is equal to the asymptotic risk of ridge regression, an explicit regularization procedure.  Here we review the setup utilized in \cite{Lejeune2019} and demonstrate that the same procedure applied to an augmented design is equivalent to one in which more shrinkage is applied to the original data.

Assume now that we have data of the form $\bm{Z}_1, ..., \bm{Z}_n$ where each $\bm{Z}_i = (\bm{X}_i,Y_i)$ and
\begin{align*}
	Y_i = \bm{X}_i'\beta + \epsilon_i 
\end{align*} 
where $Y_i \in \mathbb{R}$ denotes the response, the features $\bm{X}_i \in \mathbb{R}^{p}$  are drawn i.i.d.\ from $\mathcal{N}_p(0_{p \times 1}, \Sigma)$, and the $\epsilon_i$ are i.i.d.\ with mean 0 and variance $\sigma_\epsilon^2$ and are independent of $\bm{X}$. 

To build OLS ensembles, we draw $B$ submatrices by applying row subsampling to the observations and column subsampling on $\bm{X} = [\bm{X}_1, \dots, \bm{X}_n]'$. Let $S_b$ and $T_b$ denote the sets of column and row indices, respectively, selected in the $b^{th}$ model, while $\bm{S}_b$ and $\bm{T}_b$ denote the subsampling matrices obtained by selecting the the columns from $I_p$ and $I_n$ corresponding to the indices in $S_b$ and $T_b$.  Let $\mathcal{S}$ and $\mathcal{T}$ denote the entire collections of all possible $S_b$ and $T_b$, respectively. For each base learner, the OLS minimum-norm estimator is given by
\begin{align*}
	\hat{\beta}^{(b)} = \bm{S}_b\left( \bm{T}_b' \bm{X} \bm{S}_b\right)^+ \bm{T}_b Y
\end{align*}
where $(\cdot)^+$ denotes the Moore-Penrose pseudoinverse, so that the estimated coefficients of the ensemble are thus given by 
\begin{align*}
	\hat{\beta}^{ens} = \frac{1}{B} \sum_{b=1}^B \bm{S}_b\left( \bm{T}_b' \bm{X} \bm{S}_b\right)^+ \bm{T}_b Y .
\end{align*}
The risk of $\hat{\beta}^{ens}$ 
\begin{align*}
	R(\hat{\beta}^{ens}) \overset{\Delta}{=} \mathbb{E}_{\bm{x}}\left[ \left\langle \bm{x}, \beta - \hat{\beta}^{ens} \right\rangle \right] 
	= \left\langle \beta - \hat{\beta}^{ens}, \Sigma(\beta - \hat{\beta}^{ens}) \right\rangle
\end{align*}
is defined as the expected squared error at an independent point $\bm{x}$ and by the independence between $\bm{X}$ and $\epsilon$, \cite{Lejeune2019} show that the expected risk over $\epsilon$ can be decomposed into bias and variance terms as
\begin{align*}
	\mathbb{E}_{\epsilon} \left[R(\hat{\beta}^{ens})\right] &= \text{bias}(\hat{\beta}^{ens}) + \text{variance}(\hat{\beta}^{ens})= \frac{1}{B^2} \sum_{b, c = 1}^B \text{bias}_{b,c}(\hat{\beta}^{ens}) + \frac{1}{B^2} \sum_{b, c = 1}^B \text{var}_{b,c}( \hat{\beta}^{ens})	
\end{align*}	
where 
\begin{align*}	
	\text{bias}_{b,c}(\hat{\beta}^{ens}) &=  \left\langle \beta \beta', \ \left(I_p - \bm{S}_b\left(\bm{T}_b'\bm{X}\bm{S}_b\right)^+\bm{T}_b'\bm{X}\right)'\Sigma \left(I_p - \bm{S}_c\left(\bm{T}_c'\bm{X}\bm{S}_c\right)^+\bm{T}_c'\bm{X}\right) \right\rangle	\\
	\text{var}_{b,c}(\hat{\beta}^{ens}) &=  \sigma^2 \left\langle \bm{S}_b\left(\bm{T}_b'\bm{X}\bm{S}_b\right)^+\bm{T}_b', \ \Sigma \bm{S}_c\left(\bm{T}_c'\bm{X}\bm{S}_c\right)^+\bm{T}_c' \right\rangle.	
\end{align*}

\noindent The authors then employ the following assumptions to allow for a more precise evaluation of the risk. 

\begin{assumption}\label{assumption: finite_subsample}(Finite Subsampling)
	The subsets in the collections $\mathcal{S}$ and $\mathcal{T}$ are selected at random such that $|S_b| < |T_b| -1$ and that the following hold:
\begin{enumerate}
	\item $Pr(j\in S_b) = \frac{|S_b|}{p}$ for all $j \in [p] = \{1,2,\dots,p\}$
	\item $Pr(m\in T_b) = \frac{|T_b|}{n}$ for all $m \in [n]$
	\item The subsets $S_1,S_2,\dots,S_B,T_1,\dots,T_B$ are conditionally independent given the row subsample sizes $(|T_b|)_{b=1}^B$.
\end{enumerate} 
\end{assumption}

\begin{assumption}\label{assumption: asymptotic_subsample}(Asymptotic Subsampling)
	For some $\alpha, \eta \in [0,1]$, the subsets in the collections $\mathcal{S}$ and $\mathcal{T}$ are selected randomly such $|S_b|/p \overset{a.s.}{\longrightarrow} \alpha$ as $p\rightarrow \infty$ and  $|T_b|/n \overset{a.s.}{\longrightarrow} \eta$ as $n \rightarrow \infty$ for all $b \in [B]$.
\end{assumption}

Furthermore, it is assumed that $\Sigma = I_p$, that $\|\beta\|_2 = 1$, and that $p/n \rightarrow \gamma$ with $\eta > \alpha\gamma$ as $n,p \rightarrow \infty$.

Under these assumptions, conditional on the subset sizes, the expected risk of the bias and variance components over $\bm{X}$, $\mathcal{S}$ and $\mathcal{T}$ converge almost surely as follows:
\begin{align*}
	\mathbb{E}_{\bm{X},\mathcal{S},\mathcal{T}}\left[ \text{bias}_{b, c} (\hat{\beta}^{ens})\right] &\overset{a.s.}{\rightarrow} B(\alpha):= \left\lbrace 
	\begin{matrix}
	(1-\alpha)^2\left( 1 + \frac{\alpha^2\gamma}{1 - \alpha^2\gamma}\right)	& \text{if }b \neq c	\\
	(1-\alpha)\left( 1 + \frac{\alpha\gamma}{\eta - \alpha\gamma}\right)	& \text{if }b = c\\
	\end{matrix}
	\right.  \\
	\mathbb{E}_{\bm{X},\mathcal{S},\mathcal{T}}\left[ \text{var}_{b, c} (\hat{\beta}^{ens}) \right] &\overset{a.s.}{\rightarrow} V(\alpha):= \left\lbrace 
	\begin{matrix}
	\frac{\sigma^2\alpha^2\gamma}{1 - \alpha^2\gamma} 	& \text{if }b \neq c	\\
	\frac{\sigma^2\alpha\gamma}{\eta - \alpha\gamma}	& \text{if }b = c . \\
	\end{matrix}
	\right.  
\end{align*}

\noindent Thus, for an OLS ensemble built with $\alpha = \alpha_1$ with subsamples drawn such that $|S_b| = \lfloor \alpha_1 p \rfloor$ and $|T_b| = \lfloor \eta_1 n \rfloor$, $\mathbb{E}_{\bm{X},\mathcal{S},\mathcal{T}}\left[ \text{bias}_{b, c} (\hat{\beta}^{ens})\right]$ and $\mathbb{E}_{\bm{X},\mathcal{S},\mathcal{T}}\left[ \text{var}_{b, c} (\hat{\beta}^{ens})\right]$ will converge almost surely to $B(\alpha_1)$ and $V(\alpha_1)$ respectively. 

Now suppose that the same kind of subsampled OLS ensemble is constructed on an augmented feature space where $\bm{X}$ is augmented with $\bm{N} = [\bm{N}_1, \dots, \bm{N}_n]' \in \mathbb{R}^{n \times q}$, and where the $\bm{N}_i$ are drawn i.i.d.\ from $N_q(0_{q \times 1},I_q)$. Let $S_b^\star$ and $T_b^\star$ denote the subsampling indices on the $b^{th}$ model constructed on this augmented design $\left[\bm{X} \ \bm{N} \right]$ and suppose that the subsampling sizes remain the same as in the OLS ensemble constructed on the original data so that $|S_b^\star| = |S_b|$ and $|T_b^\star| = |T_b|$. Furthermore, suppose that the number of additional features $q\rightarrow\infty$ as $p\rightarrow\infty$ such that $\frac{q}{p}\rightarrow \theta$ for some constant $\theta > 0$. Under these assumptions, 
\[\frac{|S_b|}{p+q} \rightarrow \frac{\alpha_1}{1+\theta} = \alpha_1^\star < \alpha_1\]
and so $\mathbb{E}_{\bm{X},\mathcal{S},\mathcal{T}}\left[ \text{bias}_{b, c} (\hat{\beta}^{ens})\right]$ and $\mathbb{E}_{\bm{X},\mathcal{S},\mathcal{T}}\left[ \text{var}_{b, c} (\hat{\beta}^{ens})\right]$ converge to $B(\alpha_1^\star)$ and $V(\alpha_1^\star)$, respectively. Thus, constructing an OLS ensemble on an augmented design has the same effect as constructing the ensemble on the original design with a smaller subsampling rate, thereby producing a more regularized estimator.

\subsection{Implicit Regularization and Ridge Regression}
In addition to the work described above, an intriguing collection of work has emerged in recent years on the so-called ``double-descent" phenomenon coined by \cite{Belkin2019}, whereby the generalizability error of models may sometimes continue to improve beyond the point of interpolation where training error vanishes.  \cite{Hastie2019} followed up this work with an impressive and thorough analysis on the behavior of minimum norm interpolation for high-dimensional least squares estimators.  While this work focused on the ``ridgeless" setting, interesting related results have also been established for ridge and kernel ridge regression.  \cite{kobak2019optimal} showed that for a standard ridge estimator of the form 
\[
\hat{\beta}_\lambda = (\bm{X}' \bm{X} + \lambda I)^{-1} \bm{X}' Y
\]
\noindent the optimal penalty $\lambda$ can be 0 or negative even when $p \gg n$.  In particular, this may happen when the majority of signal comes from a small subset of high-variance features due to an implicit regularization effect offered by a larger collection of relatively low-variance noise features.  In very recent work, \cite{jacot2020implicit} consider ridge estimators acting on a (possibly larger) transformed feature space consisting of Gaussian random features and show that such an estimator with ridge penalty $\lambda$ is close to a kernel ridge regression estimator with effective penalty $\tilde{\lambda}$ where $\tilde{\lambda}>\lambda$.  \cite{dAscoli2020} consider a similar random feature setup in investigating the double descent behavior of neural networks and provide a thorough review of much of the recent work on interpolation where we would refer interested readers.  

In motivating the AugBagg procedure proposed above, we turn to a key result from \cite{kobak2019optimal}.  As above, assume we have (original) training data of the form $(\bm{X},Y)$ where $y = \bm{x}' \beta + \epsilon$ and let $\hat{\beta}_{\lambda}$ denote the ridge estimator of $\beta \in \mathbb{R}^{p}$.  Now consider a new estimator $\hat{\beta}_q$ formed by performing minimum norm least squares and taking only the first $p$ elements after augmenting $\bm{X}$ with $q$ additional i.i.d.\ noise features, each with mean $0$ and variance $\lambda/q$.  The theorem below shows that augmenting the original design with low-variance noise features produces an equivalent regularization effect to ridge regression.

\begin{theorem}\label{thm_ridge}
	\emph{[\cite{kobak2019optimal}]} Under the setup described above,
	\[\hat{\beta}_q \xrightarrow[q\rightarrow \infty]{a.s.} \hat{\beta}_\lambda.\]
	Furthermore, for any $\bm{x}$, let $\hat{y}_\lambda = \bm{x}'\hat{\beta}_{\lambda}$ denote the ridge prediction and let $\hat{y}_{Aug}$ be the prediction generated by the augmented model that includes the additional $q$ parameters using $\bm{x}$ extended with $q$ random elements generated in the same fashion. Then
	\[\hat{y}_{Aug} \xrightarrow[q\rightarrow\infty]{a.s.} \hat{y}_\lambda.\] 
\end{theorem}

\begin{figure}[t!]
	\centering
	\includegraphics[width=0.49\linewidth]{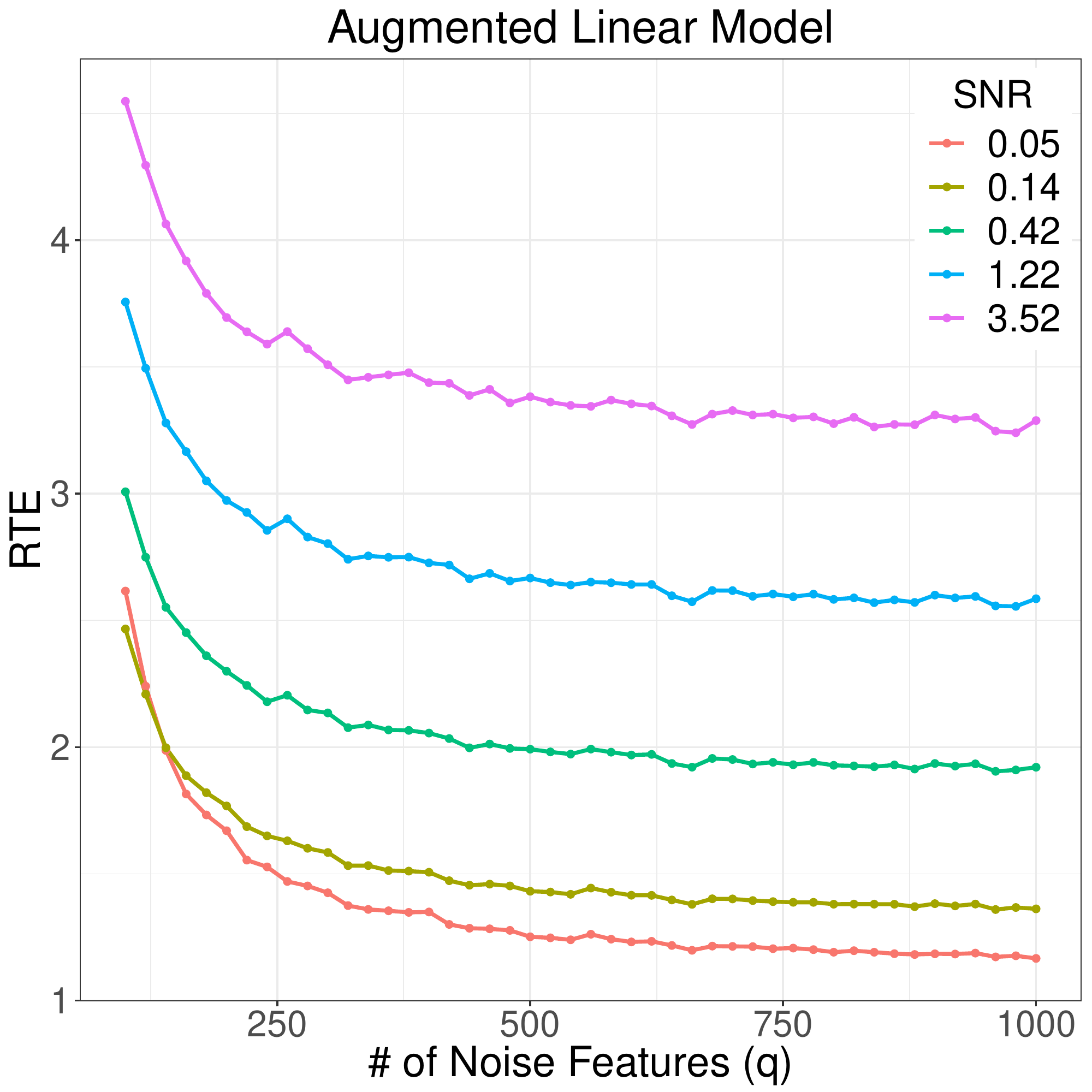}
	\caption{Performance of augmented linear model across different SNRs as increasingly many noise features are added to the model.}
	\label{fig:AugLin}
\end{figure}

\cite{kobak2019optimal} go on to note that a direct but surprising consequence of this result is that ``\emph{adding random predictors with some fixed small variance could in principle be used as an arguably bizarre but viable regularization strategy similar to ridge regression}."  Furthermore, the final statement in Theorem \ref{thm_ridge} implies that the expected MSE of the augmented model (i.e.\ the non-truncated model that includes the $q$ additional noise features) converges to the MSE of the ridge estimator as $q \rightarrow \infty$.  In particular, note that when the optimal $\lambda$ is non-zero, the augmented model with noise features generated according to the procedure outlined above will outperform the model that utilizes only the original data.

Figure \ref{fig:AugLin} gives a demonstration of this surprising result.  Here we utilize the same linear model setup described in previous sections with $n=100$ observations, $p=75$ features, the first $s=5$ of which are signal with a coefficient equal to 1.  For each SNR, we begin by generating 100 independent datasets and perform cross-validation on each to obtain 100 estimates of the optimal value of $\lambda$; the final estimate $\hat{\lambda}_{opt}$ is taken as the median across these.  Then, for each combination of SNR and $q$, we generate an independent training set where the $q$ additional noise features are sampled i.i.d.\ from $\mathcal{N}(0,\hat{\lambda}_{opt}/q)$.  The minimum-norm OLS estimator is then calculated via the singular value decomposition and the relative test error is recorded on an independent test set with 100 observations.  The entire process is repeated 100 times and the average relative test error is shown in Figure \ref{fig:AugLin}.  In each case, we see clearly that the model error decreases as more noise features are added into the model.

Suppose now that we build ensembles of estimators of the kind in Theorem \ref{thm_ridge} by drawing $B$ subsamples, constructing the estimators on each subsample, and averaging.  Similar to the setup used above in \cite{Lejeune2019}, let $T_b \subseteq [n]$ be the set of indices of selected observations in the $b^{th}$ subsample and let $\bm{T}_b$ be the $n\times |T_b|$ matrix obtained by selecting columns from $I_n$ corresponding to the indices in $T_b$. Construct $\hat{\beta}_q^{(b)}$ as above based on $\bm{T}_b' \bm{X}$ and $\bm{T}'_bY$, which denote the design matrix and response, respectively, corresponding to the observations selected in $b^{th}$ subsample. 
The final ensemble coefficient estimate formed by averaging the augmented minimum norm estimators is given by 
\[\hat{\beta}^{ens} = \frac{1}{B}\sum_{b=1}^B \hat{\beta}_q^{(b)}\]
where, by Theorem \ref{thm_ridge},
\[ \hat{\beta}^{ens} = \frac{1}{B}\sum_{b=1}^B \hat{\beta}_q^{(b)} \xrightarrow[q\rightarrow \infty]{a.s.} \frac{1}{B}\sum_{b=1}^B \hat{\beta}_\lambda^{(b)} \]
with 
\begin{align*}
	\hat{\beta}_\lambda^{(b)} &= (\bm{X}'\bm{T}_b\bm{T}_b'\bm{X} + \lambda I_p)^{-1}\bm{X}'\bm{T}_b\bm{T}_b'Y.
\end{align*}

Now consider a simple setting where $p=n$ and $\bm{X} = I_n$ and let $\eta$ denote the subsampling rate so that $|T_b|/n \rightarrow \eta \in (0,1]$. Let $C$ be a $n \times n$ diagonal matrix where $C_{ii}$ is the number of times that the $i^{th}$ observation appears in the $B$ subsamples and let $\lambda_q = \frac{1 + \lambda -\eta}{\eta} \geq \lambda $. Then we have
\begin{align*}
	\hat{\beta}^{ens} \, \xrightarrow[q\rightarrow \infty]{a.s.} \, \frac{1}{B}\sum_{b=1}^B \hat{\beta}_\lambda^{(b)} &=\frac{1}{B}\sum_{b=1}^B(\bm{T}_b\bm{T}_b' + \lambda I_p)^{-1}\bm{T}_b\bm{T}_b'Y	\\
	&= \frac{1}{1 + \lambda} \frac{1}{B} CY \\
	&\overset{B \rightarrow \infty}{\longrightarrow} \frac{\eta}{1 + \lambda} Y \\
	&= \frac{1}{1 + \lambda_q} Y \\
	&= \hat{\beta}_{\lambda_q}.
\end{align*}

Thus, in this simple case, an ensemble of minimum-norm least squares estimators constructed on an augmented design produces an estimate equivalent to one produced via ridge regression on the original design.  Furthermore, the shrinkage produced by the ensemble is stronger than that of each individual base model.

\section{Implications for Variable Importance}
\label{sec:varimp}
Within any kind of black-box supervised learning framework, establishing a valid means of measuring the importance of features is of utmost importance.  Indeed, in such non-parametric regimes where model fit and behavior remain largely hidden from view, understanding how features contribute information to the prediction is paramount for scientists and practitioners.  In the context of bagging and random forests specifically, Breiman's original out-of-bagg (oob) \citep{Breiman2001} importance scores are one such popular measure, though many issues such as a preference for correlated features and those with many categories have been noted in the years following their introduction \citep{Strobl2007,Nicodemus2010,Tolocsi2011}.


Recently, however, \cite{Mentch2016} proposed a formal hypothesis testing procedure for measuring feature importance in random forests.  Given a generic relationship of the form $y = f(\bm{x}) + \epsilon$, the authors consider partitioning the original set of features $\bm{X}$ into two groups, $\bm{X}_0$ and $\bm{X}_{\text{test}}$, where the latter group contains the features of interest so that a null hypothesis of the form
\begin{equation}
H_0: \, f(\bm{X}_0,\bm{X}_{\text{test}}) = f_0(\bm{X}_0)
\label{eqn:h0}
\end{equation}
\noindent may be rejected whenever the features in $\bm{X}_{\text{test}}$ make a significant contribution to predicting the response.  The authors propose to evaluate the hypothesis in (\ref{eqn:h0}) by constructing two separate random forest models:  one constructed on the original data and one constructed on an altered dataset where the features in $\bm{X}_{\text{test}}$ are either substituted for randomized replacements independent of the response or dropped from the model entirely.  Predictions from each forest are then computed at a number of test points and the differences are combined to form an appropriate test statistic.  \cite{Coleman2019} recently proposed a more computationally efficient and scalable permutation test alternative that involves exchanging trees between the two forests.  

Crucially, the tests described above rely on measuring the difference between either raw predictions or predictive accuracy between two tree-based ensembles constructed on different training sets.  Both papers advocate for replacing the features under investigation with randomized alternatives, noting that the tests can potentially produce spurious results when features are instead dropped from the second model, though neither provides a detailed explanation as to why this occurs.  Elsewhere in the literature, alternative tests specifically propose to evaluate feature importance by measuring the drop in performance when the features in question are removed from the model.  Such is the case, for example, with the \textbf{L}eave-\textbf{O}ut-\textbf{Co}variates (LOCO) measure proposed by \cite{Lei2018} in the context of conformal inference and most recently in the tests proposed by \cite{Williamson2020}.  Furthermore, though often done informally, it remains common throughout the broader scientific literature for authors to argue for the importance of particular variables based on decreases in model performance when such variables are excluded.

The results presented in the sections above present a substantial concern with such measures.  In particular, if model performance can be improved simply by adding randomly generated features that are (at least conditionally) independent of the response, then observing a significant improvement in accuracy when a particular set of features is included does not imply that any relationship to the response or even the other covariates need exist.  To emphasize this point, we implement the test for variable importance recently developed in \cite{Williamson2020} and investigate its behavior under simulated settings similar to those used in the previous sections.  

We utilize the same linear model setup as in previous sections with $p=5$ original signal features sampled from $\mathcal{N}_p(\bm{0}, \Sigma)$ with $\Sigma_{ij} = \rho^{|i-j|}$ and $\rho = 0.35$ and consider adding $q$ additional noise features to test for importance.  These noise features are either independent of the original five features or are correlated with a randomly selected signal feature with correlation strength $r$.  Thus, relative to the sort of generic null hypothesis specified in (\ref{eqn:h0}), our default set of features consist of the original signals so that $\bm{X}_0 = (X_1, ..., X_5)$ and the features under investigation are those additional noise features being added, $\bm{X}_{\text{test}} = (N_1, ..., N_q)$.  As done previously, the error in the model is adjusted to produce a pre-specified SNR.

To carry out the procedure in \cite{Williamson2020}, for each test, we create a training set $\mathcal{D}_n$ with $n=500$ observations and two test sets $\mathcal{D}_{Test, 1}$ and $\mathcal{D}_{Test, 2}$ with $n_1 = n_2 = 1000$ observations.  Let $\bm{X}$ denote the original $n \times (p+q)$ design matrix and $\bm{X}^*$ denote the design matrix where the $q$ noise features of interest are either dropped or replaced with a random substitute.  Thus, for ``drop tests", $\bm{X}^*$ will be of dimension $n \times p$ whereas for ``replacement tests", $\bm{X}^*$ will be of dimension $n \times (p+q)$.  We construct two decision tree ensembles, each with 500 trees.  The first ensemble is created by by performing bagging on the original data $(Y,\bm{X})$ and the second by bagging on the modified data $(Y,\bm{X}^*)$.  We then record the test error of the first ensemble on $\mathcal{D}_{Test, 1}$ and the test error of the second, modified ensemble on $\mathcal{D}_{Test, 2}$.  The difference in test MSEs between the ensembles is then given by 

\[T = \frac{1}{n_{1}} \sum_{i \in \mathcal{D}_{Test,1}} (Y_i - \hat{f}(\bm{X}_i))^2  \; - \; \frac{1}{n_{2}} \sum_{i \in \mathcal{D}_{Test,2}} (Y_i - \hat{f}^*(\bm{X}_i^*))^2 . \]

\noindent \cite{Williamson2020} show that $T \approx \mathcal{N}(\Delta, \sigma^2)$ where
\[\Delta = \mathbb{E}\left[ (Y - \mu(\bm{X}))^2 \right] - \mathbb{E}\left[ (Y - \mu^*(\bm{X}^*))^2 \right] \]
\noindent and the variance $\sigma^2$ can be estimated with
\[\hat{\sigma}^2 = \frac{s_1^2}{n_{1}} + \frac{s_2^2}{n_{2}}\]
where $s_1^2$ and $s_2^2$ are the empirical variances of $(Y_1 - \hat{f}(\bm{X}_1))^2, ..., (Y_{n_1} - \hat{f}(\bm{X}_{n_1}))^2$ and $(Y_1 - \hat{f}^*(\bm{X}^{*}_{1}))^2, ..., (Y_{n_2} - \hat{f}^*(\bm{X}^{*}_{n_2}))^2$, respectively.  Since we would like to claim that the features of interest are important whenever the test MSE is larger in the second ensemble where those features are either dropped or replaced, the null and alternative hypotheses of interest are $H_0: \Delta = 0$ and $H_1: \Delta < 0$, respectively.


\begin{figure}[t!]
	\centering
	\includegraphics[width=0.49\linewidth]{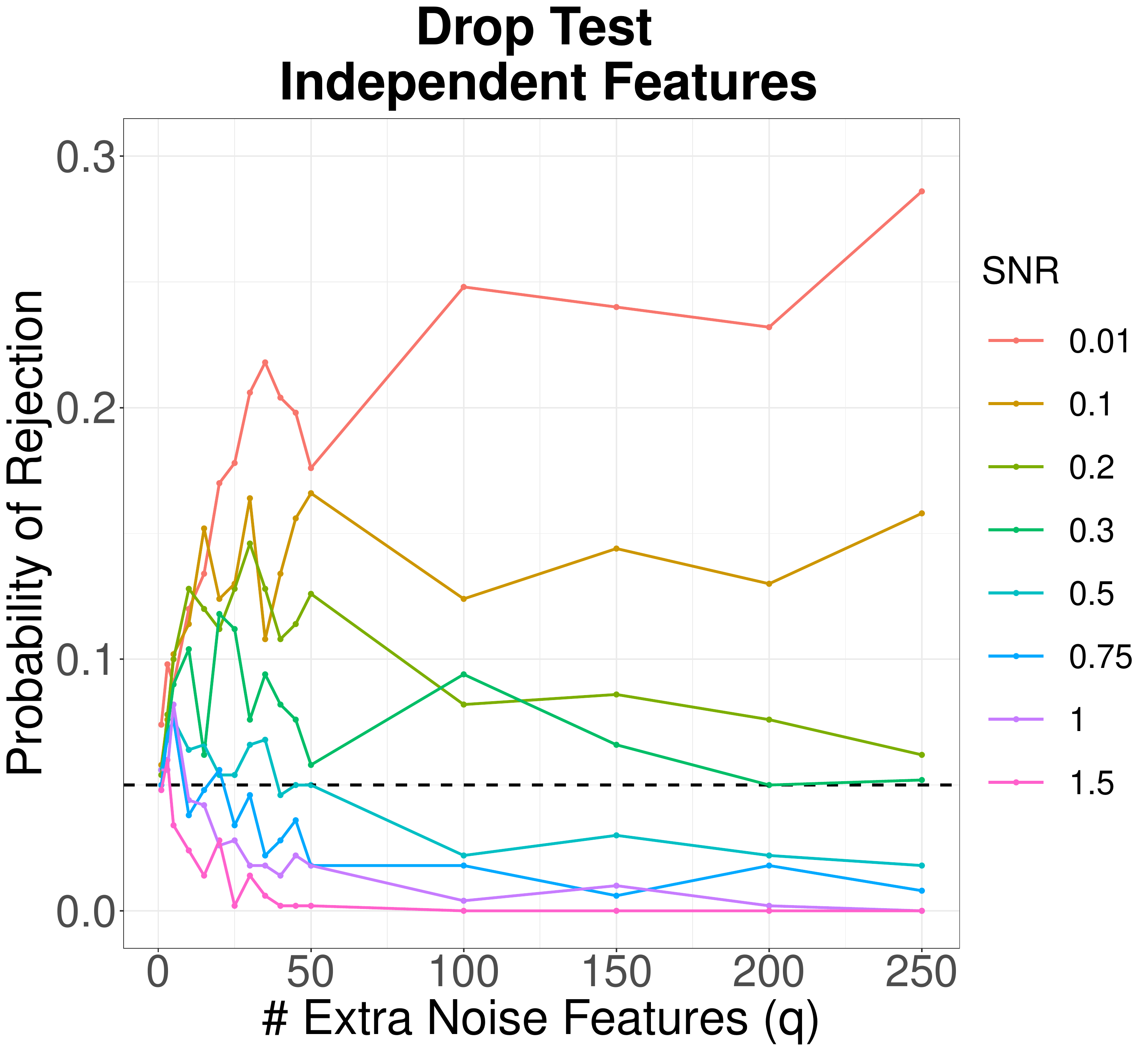}
	\includegraphics[width=0.49\linewidth]{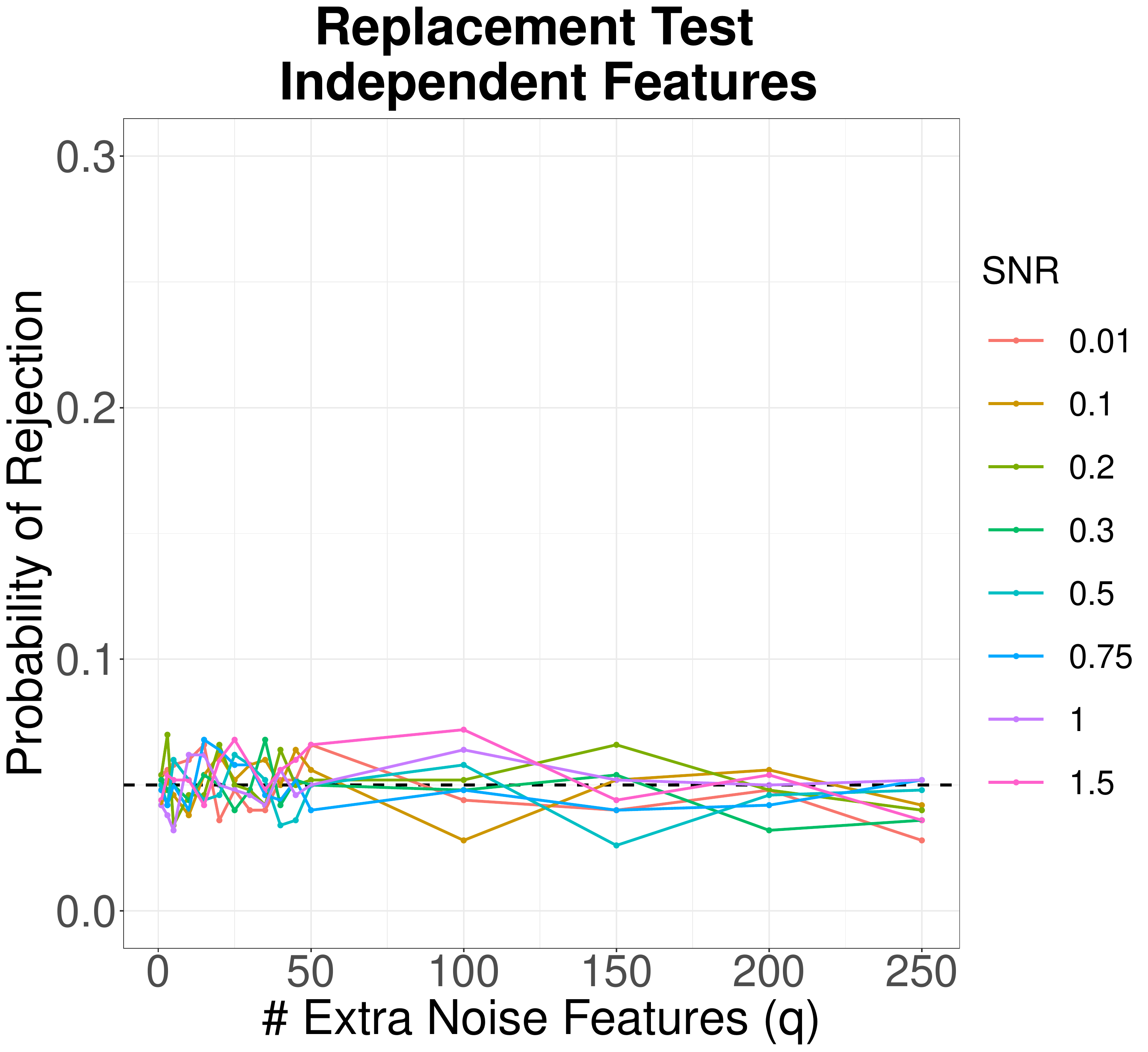} \\
	\vspace{2mm}
	\includegraphics[width=0.49\linewidth]{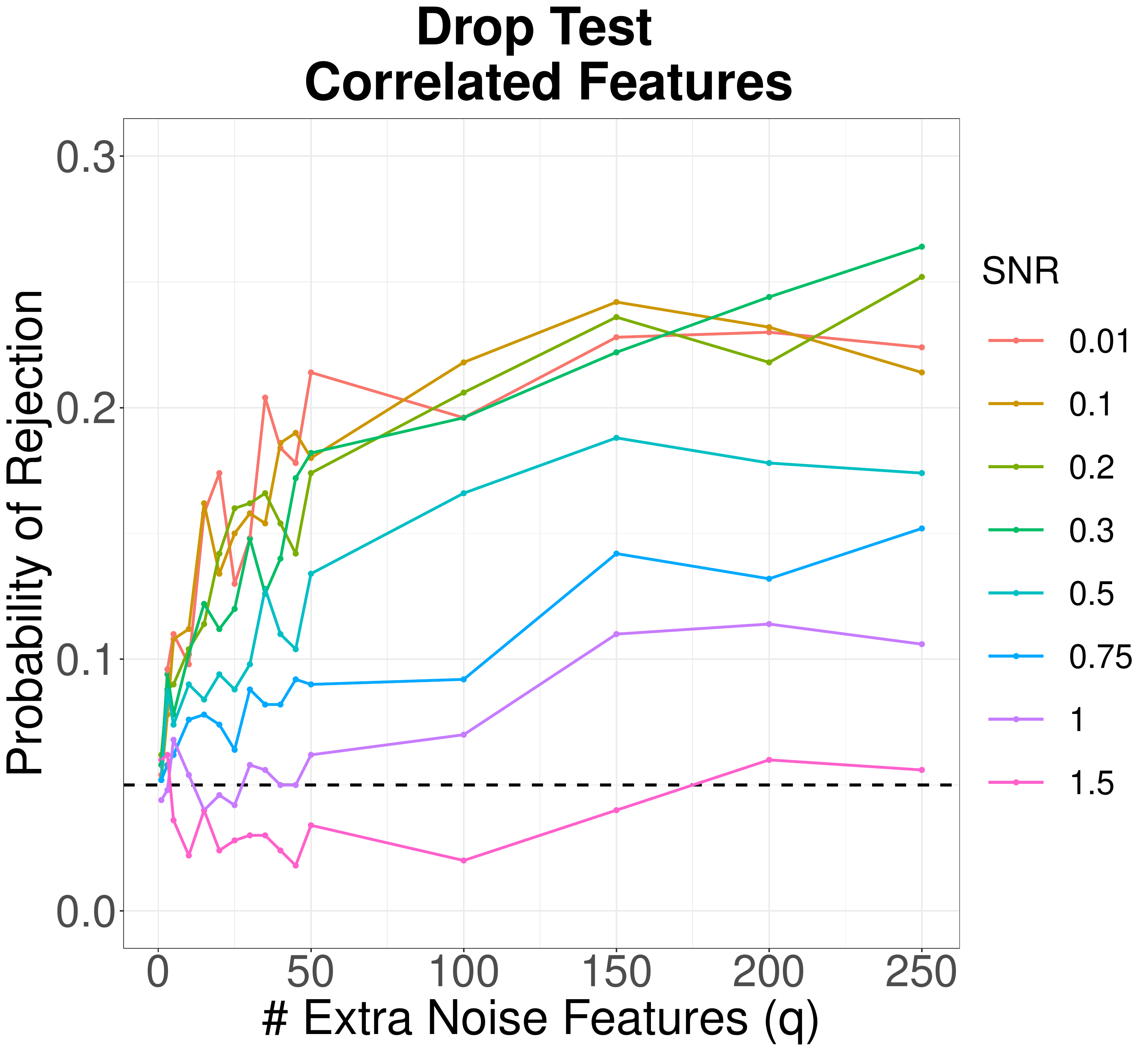}
	\includegraphics[width=0.49\linewidth]{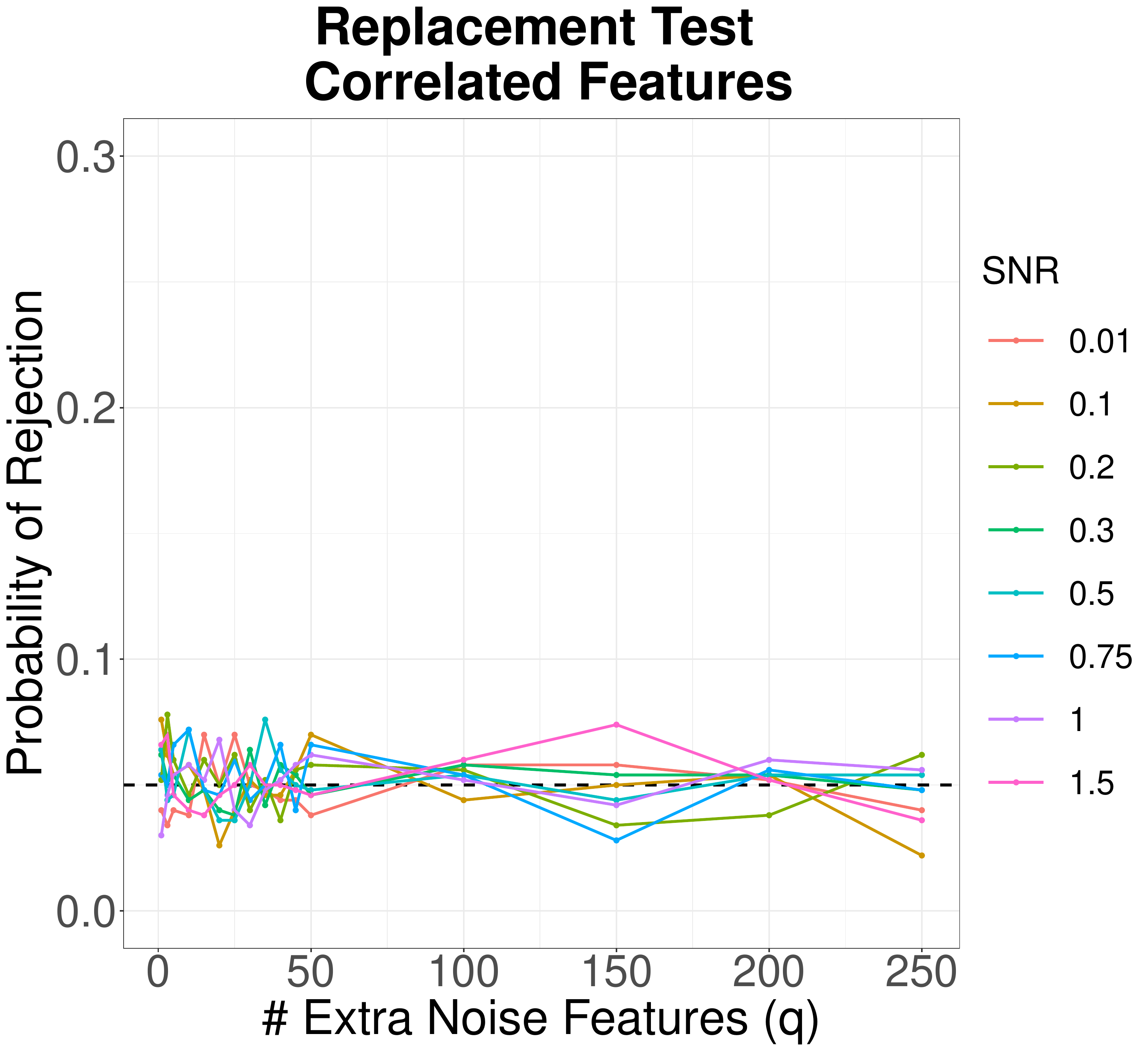}
	\caption{Probability of rejecting the null hypothesis and concluding an additional independent set of noise features are important when dropping the features in question (left column) vs replacing the features in question (right column) when those features are independent (top row) vs correlated (bottom row).}
	\label{fig:tests}
\end{figure}

Figure \ref{fig:tests} shows the probability of rejecting $H_0$ and concluding the additional noise features are important across various SNRs and numbers of additional features when those features are either dropped or replaced by null substitutes.  For these as well as each of the tests deployed below, we set the nominal level to the standard $\alpha = 0.05$ so that if the tests are performing as intuitively expected, we should only see the null hypothesis to be rejected (indicating that the noise features are significant) about 5\% of the time.  However, it is immediately seen that whenever the additional noise features are independent of the original features (Figure \ref{fig:tests} Top Left), the tests that drop the features under investigation reject nearly 30\% of the time at low SNRs when a large number of features are being tested.  When those additional noise features are generated in a correlated fashion (Figure \ref{fig:tests} Bottom Left), the rejection probabilities for drop tests rise above the nominal level of 0.05 even for larger SNRs as the number of noise features increases.


Though troubling, these results are not at all surprising given the empirical results in Section \ref{sec:sims} that showed strong evidence of improved performance when additional noise features are added to the model.  These tests simply make clear that such improvements are routinely large enough to register statistical significance.  We caution readers from drawing too much from the particular rejection probabilities shown in the left column of Figure \ref{fig:tests}.  These empirical results should in no way be seen as guidelines for how often or under what settings such tests will produce inflated rejection proportions.  Rather, the amount by which these kinds of tests inflate the anticipated rejection proportion will depend entirely on the relationships within the data as well as the power of the particular testing procedure employed.  Indeed, similar testing procedures with higher power would reject even more often than shown in Figure \ref{fig:tests} for the same datasets.

Tests such as those proposed in \cite{Mentch2016} and \cite{Coleman2019}, claim to offer a more robust testing procedure whenever the features under investigation are replaced by randomly generated substitutes rather than being dropped from the model entirely.  And indeed, from the right column of Figure \ref{fig:tests} it is readily observed that regardless of the SNR or the dependence structure of the noise features on the original features, these replacement tests maintain a rejection rate very near the nominal rate of 5\%.

\begin{figure}[t!]
	\centering
	\includegraphics[width=0.49\linewidth]{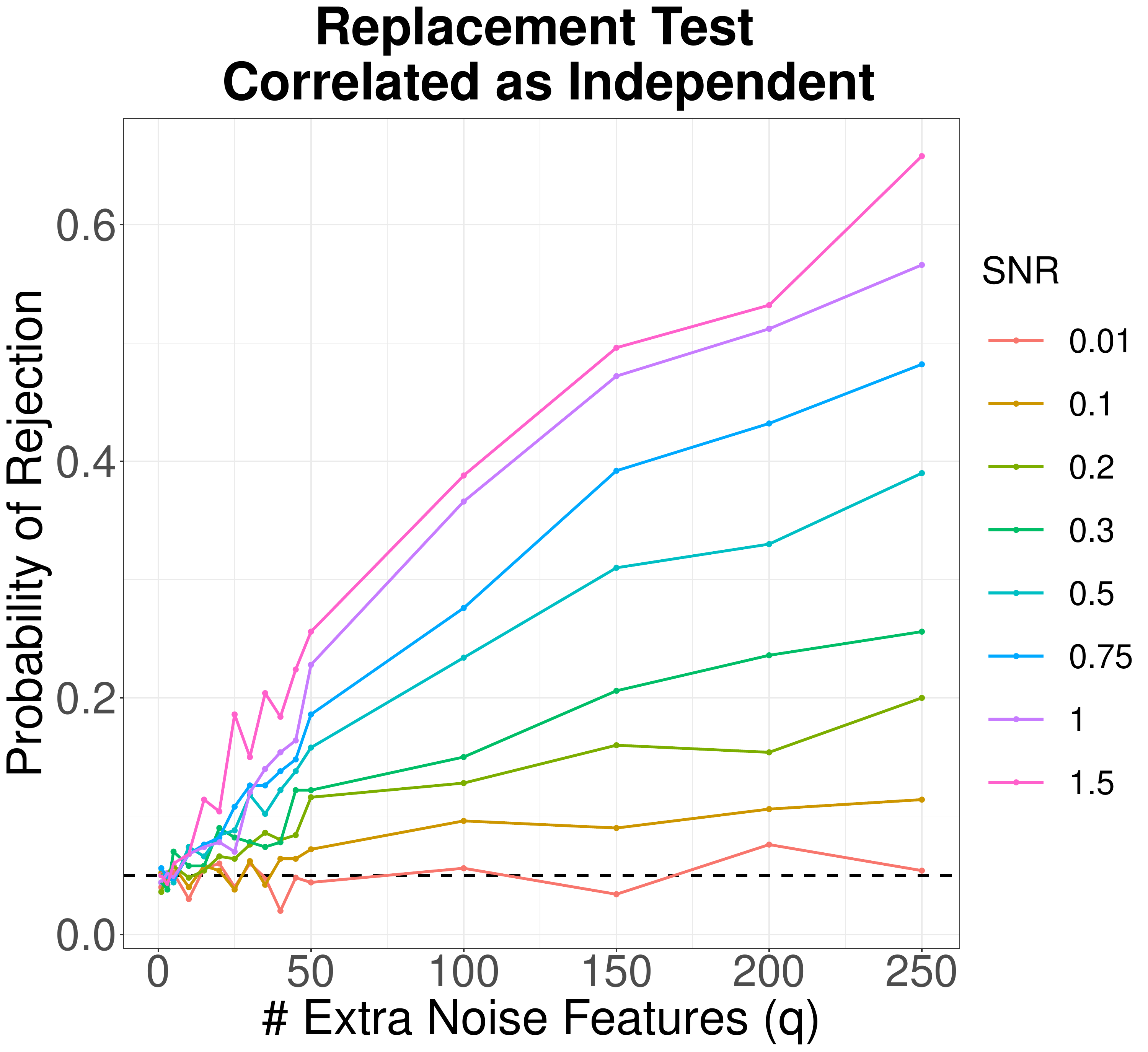}
	\includegraphics[width=0.49\linewidth]{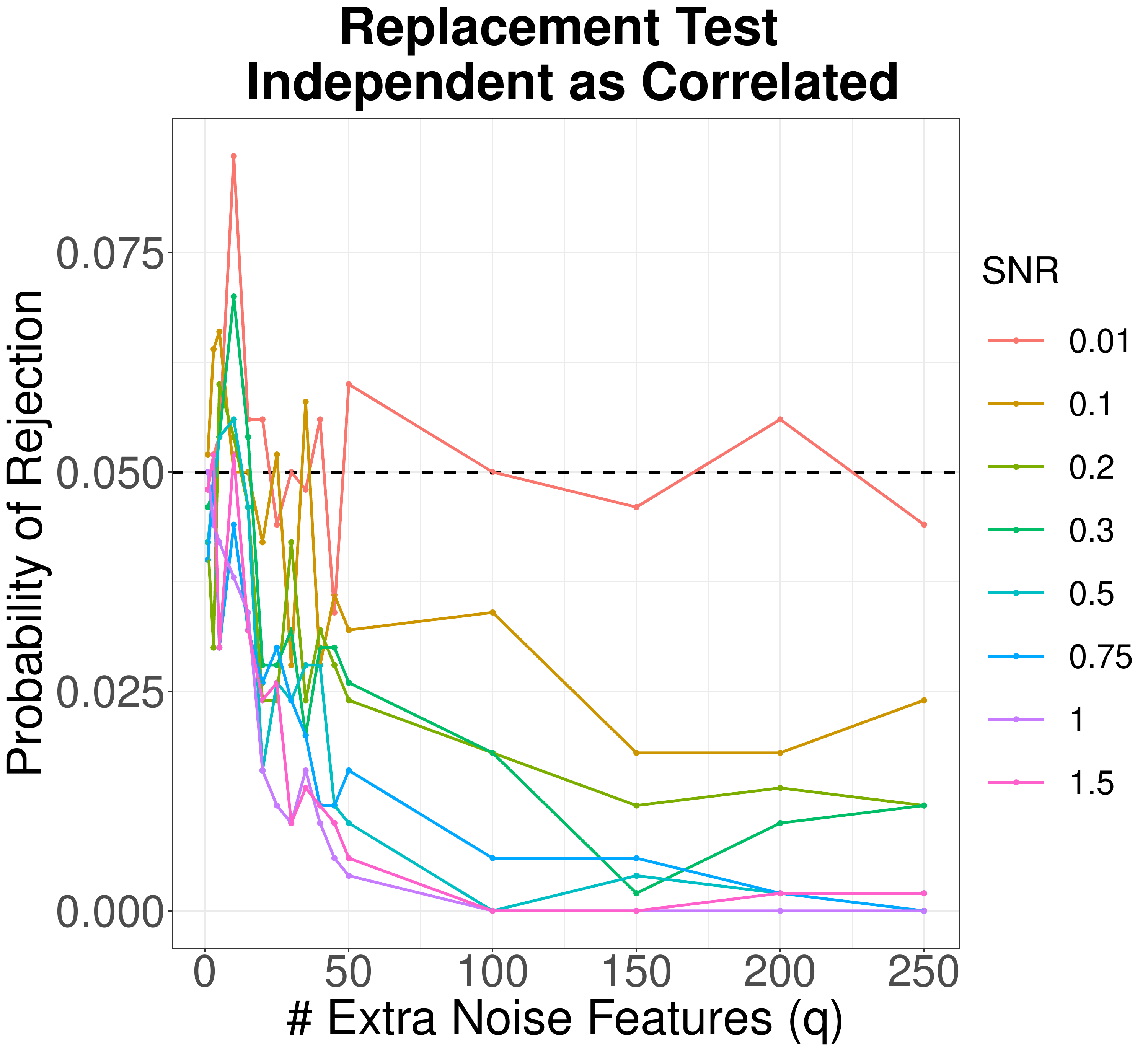}
	\caption{Probability of rejecting the null hypothesis using replacement tests where correlated features are replaced with independent features (left) and where independent features are replaced with correlated features (right).}
	\label{fig:testscomp}
\end{figure}


Unfortunately, carrying out accurate replacement-style tests in practice is easier said than done.  In the plots shown in the right-hand column of Figure \ref{fig:tests}, the replacement noise features are sampled from exactly the same distribution as the original noise features being tested for importance.  In practice, of course, the distribution of the features in question is unknown.  Figure \ref{fig:testscomp} compares the performance of these replacement tests whenever noise features of one kind are replaced by noise features of another kind.  On the left, the original noise features are randomly correlated with an original signal feature at $r=0.7$ and these features are replaced with independent noise features.  Here we again notice a troubling trend:  the test has a very high probability of rejecting across all but the lowest SNRs and this probability appears to increase with $q$.  Perhaps even worse is the fact that the rejection probabilities appear to be increasing at a faster rate at higher SNRs.  Thus, even in ``good data" settings, it appears that such tests are very likely to cause correlated noise features to appear important whenever testing against the performance of a model using only independent noise (or, for example, permutations of the original features) as a substitute.  

While this setting is likely most representative of what might often happen in practice, for completeness, we also consider the opposite setting in the plot on the right where independent noise features are replaced with ones correlated with a randomly selected feature in $\bm{X}_0$. Here we see the opposite trend:  the rejection probabilities appear small for all but the lowest SNRs and also appear to be shrinking as $q$ grows.

\subsection{Bad Tests or Bad Interpretations?}
\label{subsec:testdisc}
Given these results, one may be tempted to conclude that tests of the style proposed in work such as \cite{Mentch2016}, \cite{Lei2018}, \cite{Coleman2019}, and \cite{Williamson2020} are simply ``bad" because the outcomes are ``wrong" far too often.  Indeed, if rejecting the null hypothesis in these types of procedures is taken to mean that the features in question are ``important" and ``important" is taken to mean that those features possess some unique explanatory power for the response not captured by the other features available, then certainly such tests would appear to be highly problematic as the rejection rates in the above settings very often lie far above the nominal level of $\alpha=0.05$.

In our view, however, such an understanding is too naive.  The demonstrations above do not necessarily imply that anything is wrong with the tests themselves.  Rejecting the null hypotheses in such tests means only that there is evidence that the features in question improve model performance when included.  The simulations in Section \ref{sec:sims}, however, suggest that even the inclusion of additional noise features can improve model performance, sometimes to a dramatic degree.  

This situation highlights the crucial need for precise language in discussions of feature importance.  While ``predictive improvement" intuitively feels like a natural proxy, it seems quite unlikely that features independent of the response (at least conditionally) ought to ever be considered ``important" for most practical purposes.  Certainly this is the case whenever scientists argue that particular features must be collected in order to construct the optimal predictive model or when arguing that features generated by a new piece of technology can lead to further improved model performance over those that were previously available.

In situations such as these, it seems that what is really being sought is not a measure of how ``important" certain features may be, but rather how ``essential" they are.  Even when additional variables improve model performance, we really seek to determine whether they do so meaningfully or significantly more than randomized alternatives.  The results in the preceding section also highlight the potential issues with replacing features by randomized replacements from a different distribution and thus might suggest some promise for procedures involving knockoff variables \citep{Barber2015,Candes2018} that explicitly attempt to generate randomized replacements from the same distribution as the original copies.

\section{Discussion}
\label{sec:discussion}
The work in the preceding sections introduced the idea of augmented bagging (AugBagg), a simple procedure identical to traditional bagging except that additional noise features, conditionally independent of the response, are first added to the feature space.  Surprisingly, we showed that this simple modification to bagging can lead to drastic improvements in model performance, sometimes even outperforming well-established alternatives like an optimally-tuned random forest.  Performance gains appear most dramatic at low SNRs, though the introduction of correlated noise features can continue to improve performance even at higher SNRs.  The fact that performance can sometimes be dramatically improved by simply adding conditionally-independent features into the model has important implications for variable importance measures and especially in interpreting the results from tests of variable importance.

While there have been several important recent studies investigating the potential regularization effects of excess noise features, we are not aware of other work specifically advocating for a procedure that augments the original data with noisy features to achieve superior predictive accuracy.  The fact that models constructed on larger and noisier feature collections are sometimes preferable would seem to run counter to much of traditional statistical thinking.  Countless procedures have been proposed in recent decades that assume $\bm{X} = (\bm{X}_{\text{Signal}},\bm{X}_{\text{Noise}})$ and attempt to uncover the subset of signal features $\bm{X}_{\text{Signal}}$ with a minimal `false positive' rate.  Indeed, many may intuitively believe that the setting where all available features are signal is something of a `gold standard' for regression.  While there may be good inferential reasons why separating signal and noise is important, this work suggests that such a task is unnecessary and perhaps even detrimental (at least for some models) whenever predictive accuracy is the primary objective.

Finally, though AugBagg may sometimes produce predictions substantially more accurate than an alternative baseline like random forests, we stress that the procedure should not be seen as replacing or superseding more efficient procedures like random forests.  As detailed in the introduction, random forests have a long documented history of off-the-shelf success and depending on the size of the data at hand, may be much more computationally feasible to implement in practice.  Indeed, while random forests reduce the number of features considered at each node, AugBagg, by construction, explicitly increases this computational burden.  Furthermore, while recent work by \cite{Mentch2019} suggests that the \texttt{mtry} parameter in random forests be tuned, moderate success can often be found at default values whereas a generic implementation of AugBagg involves tuning both the number of additional features and their correlation with the original features and we are not able to offer default values of these likely to be successful across a broad range of data settings.

\bibliographystyle{Chicago}

\end{document}